\documentclass[pdflatex,sn-basic]{sn-jnl}

\DeclareUnicodeCharacter{2003}{ }
\usepackage{graphicx}
\usepackage{multirow}
\usepackage{amsmath,amssymb,amsfonts}
\usepackage{amsthm}
\usepackage{mathrsfs}
\usepackage[title]{appendix}
\usepackage[table]{xcolor}
\usepackage{textcomp}
\usepackage{booktabs}
\usepackage{algorithm}
\usepackage{algorithmicx}
\usepackage{algpseudocode}
\usepackage{listings}
\usepackage{tabularx}
\usepackage{placeins}
\usepackage{makecell}
\usepackage{longtable}
\usepackage{array}
\usepackage{ragged2e}
\usepackage{comment}
\usepackage{pifont}
\usepackage{lmodern}
\usepackage[T1]{fontenc}
\usepackage{csquotes}
\usepackage{caption}
\DeclareUnicodeCharacter{2265}{\ensuremath{\geq}}
\providecommand{\say}[1]{\enquote{#1}}

\newcolumntype{P}[1]{>{\centering\arraybackslash}p{#1}}
\newcolumntype{L}[1]{>{\raggedright\arraybackslash}p{#1}}

\renewcommand{\arraystretch}{1.2}
\newcommand{\cmark}{\textcolor{green!60!black}{\ding{51}}}
\newcommand{\xmark}{\textcolor{red!70!black}{\ding{55}}}

\theoremstyle{thmstyleone}%
\theoremstyle{thmstyletwo}%

\theoremstyle{thmstylethree}%

\begin{document}

\title[Article Title]{The BD-LSC Dataset: Facilitating the Benchmarking of Models for Lexical Semantic Change Detection in Slang and Standard Usage}


\author*[1,2]{\fnm{Afnan} \sur{Aloraini}}
\email{afnan.aloraini@postgrad.manchester.ac.uk}
\email{A.ALOURANI@qu.edu.sa}

\author[1]{\fnm{Riza} \sur{Batista-Navarro}}
\email{riza.batista@manchester.ac.uk}

\affil[1]{
  \orgdiv{Department of Computer Science},
  \orgname{The University of Manchester},
  \orgaddress{
    \street{Oxford Road},
    \city{Manchester},
    \postcode{M13 9PL},
    \state{Greater Manchester},
    \country{United Kingdom}
  }
}

\affil[2]{
  \orgdiv{Department of Computer Science},
  \orgname{Qassim University},
  \orgaddress{
    \city{Al-Mulaida},
    \postcode{51425},
    \country{Saudi Arabia}
  }
}

\abstract{
Automatic semantic change detection aims to identify how word meanings shift over time, offering insights into both linguistic and societal change. Despite recent progress in computational lexical semantic change (LSC), existing benchmarks and methods struggle to capture bi-directional semantic change, particularly cases where words simultaneously gain and lose senses. This problem is especially challenging for words that have both slang and standard meanings. To address these gaps, we introduce two complementary benchmark datasets. The Bi-Directional Lexical Semantic Change (BD-LSC) dataset captures sense gain, sense loss, and stability across three time periods, enabling the study of complex semantic trajectories. The SlangTrack Word Sense Disambiguation (ST-WSD) dataset provides fine-grained, instance-level sense annotations for words combining slang and standard usages, supporting systematic benchmarking of WSD and semantic change detection models. Using these benchmarks, we systematically evaluate models across different methodological families: unsupervised clustering using contextualised embeddings, supervised machine learning, transformer-based models, and state-of-the-art large language models. Among the evaluated systems, the few-shot GPT-4o model achieved the strongest aggregate performance on Exact Sense Match (ESM) and multi-label accuracy; however, Macro-F1 scores near 0.5 across all systems show that rare slang senses remain difficult, which we identify as the central open challenge.
}


\maketitle

\section{Introduction}
\label{sec:bdlsc_introduction}

Lexical semantic change detection (LSCD) aims to identify how word meanings shift over time in textual data. Research on semantic change provides insights into the dynamics of language evolution and the societal, cultural, and historical factors that shape shifting patterns of usage~\citep{tahmasebi2021computational, kutuzov2018diachronic, hamilton2016cultural, xu2015computational, dubossarsky2019timeout}. Early work in this area relied on heterogeneous corpora, divergent methodologies, and inconsistent evaluation metrics, making it difficult to compare results across studies~\citep{hengchen2021challenges, dubossarsky2019timeout}. This lack of methodological consistency highlighted the need for high-quality, manually annotated diachronic corpora that support robust and reproducible semantic change evaluation~\citep{hatty2019surel}. Recent shared tasks, such as SemEval-2020 Task 1, have taken steps toward establishing common evaluation frameworks for LSC~\citep{schlechtweg2020semeval}, but existing resources remain limited in scope and linguistic coverage.

Historically, LSC has been predominantly treated as a binary classification problem, distinguishing between the presence or absence of change across two parallel time periods. However, linguistic theory shows that semantic change is often multi-directional, with word senses emerging and disappearing simultaneously~\citep{algeo1990semantic, geeraerts2009theories}. For example, the word \textit{gay} originally carried the meaning \say{joyful}, yet by the mid-twentieth century it had increasingly adopted the sense \say{homosexual}, with both meanings coexisting for several decades before the latter became dominant~\citep{robinson2012gay, li2022diachronic}. Such developments illustrate that semantic change can involve the emergence of new senses or the modification of existing ones~\citep{tahmasebi2021computational}. These semantic developments can take several forms: the introduction of a new sense to an established word, the emergence of a sense that extends an existing one, the divergence of a sense into a related but distinct meaning, or the appearance of a completely new sense unrelated to any prior usage.


Furthermore, senses can undergo transformations where the meaning of a word broadens or narrows. Broadening occurs when a word sense expands in meaning over time, while narrowing involves a word sense becoming more specific in meaning as time progresses~\citep{algeo1990semantic, geeraerts2009theories, hamilton2016cultural}. Additionally, two distinct types of change include splitting or merging word senses. In the case of merging, two word senses that previously existed independently unite over time, while splitting occurs when a single word sense divides into two distinct senses at a later point~\citep{tahmasebi2021computational, dubossarsky2019timeout}. Lastly, a sense can experience obsolescence, signifying that it is no longer in active use~\citep{bochkarev2019method}.


A comprehensive understanding of these various forms of semantic change is indispensable for researchers investigating the evolution of language across time. In this work, we reformulate LSCD to capture both directions of change between time periods. Specifically, our annotation operates at two levels: each occurrence of a target word is assigned its contextual sense (instance-level word sense disambiguation), and from the resulting per-period sense inventories we derive word-level change labels (\textit{sense gain}, \textit{sense loss}, or \textit{no change}) for each pair of time periods. We then develop both supervised and unsupervised methods to automatically detect and classify these forms of semantic change within this framework.
%

This paper tackles the critical challenge of semantic change in words that exhibit both slang and standard meanings. 
In this work, we define slang as informal, non-standard vocabulary used to signal social identity, characterised by its fluidity and rapid turnover. Slang senses evolve rapidly, particularly in online communication and across social media~\citep{dumas1978slang, hoxhaj2022using}.
 The first step in our approach involves word sense disambiguation (WSD), determining the precise meaning of a target word within a given context. This step is essential because words with slang senses often have multiple interpretations, and disambiguating them is crucial for capturing their intended meaning. 
The second step builds on this disambiguation, using it to detect and analyse different types of semantic shifts, where a word may undergo sense gain, sense loss, or experience no change across different time periods.
In other words, we not only disentangle the various senses of these words but also investigate how their meanings change over time, offering insights into the dynamic interplay between language and culture. By addressing this, the paper contributes to a deeper understanding of the complexities of slang usage and its evolving semantics. We demonstrate how both supervised and unsupervised models can be applied to WSD and LSCD in this context.


The main contributions of this research are as follows:

\begin{itemize}
\item We introduce the Bi-Directional Lexical Semantic Change (BD-LSC) dataset, the first resource designed to capture semantic change in words with both slang and standard meanings across three time periods. BD-LSC enables independent identification of sense gain, sense loss, and stability. We also release the SlangTrack Word Sense Disambiguation (ST-WSD) dataset, an instance-level benchmark providing fine-grained annotations for evaluating WSD and semantic change detection in words exhibiting slang and standard meanings.

\item We conduct a systematic evaluation of semantic change detection and word sense disambiguation across different methodological families:
(i) unsupervised clustering approaches based on contextualised embeddings,
(ii) supervised machine learning models (ML),
(iii) transformer-based language models (LMs), and
(iv) large language models (LLMs).
\item We provide detailed error analyses that highlight the strengths and limitations of current computational approaches and identify key challenges for future research.
\end{itemize}



\section{Related Work}
\label{sec:bdlsc_related_work}


\subsection{Methods for Word Sense Disambiguation (WSD)}
\label{subsec:bdlsc_wsd}

Word Sense Disambiguation (WSD) is the task of automatically determining the correct meaning of an ambiguous word based on its context, i.e., selecting the most appropriate sense of a word from a predefined inventory such as WordNet~\citep{bevilacqua2021recent, nanjundan2023analysis}. WSD methods can be broadly classified into three primary approaches: 1) Supervised Learning-Based Approach, 2) Unsupervised Learning-Based Approach, and 3) Knowledge and Rule-Based Approach.


Supervised learning approaches to WSD rely on labelled datasets such as Semantically Tagged Corpus (SemCor), Few-Shot Examples for Word Sense Disambiguation (FEWS), and WordNet for model training~\citep{bevilacqua2021recent}. Neural architectures are commonly employed to model context and disambiguate senses. For instance, stacked bidirectional Long Short-Term Memory (BiLSTM) networks with attention mechanisms have been used to generate richer sentence embeddings that capture semantic dependencies~\citep{zhang2022word}. Sense-Maintained Sentence Mixup (SMSMix) introduces data augmentation to mitigate distributional bias and better represent least frequent senses (LFS)~\citep{yoon2022smsmix}. Another example is GlossBERT, which reformulates WSD as a sentence-pair classification task, constructing context–gloss pairs with BERT models to perform disambiguation~\citep{huang2019glossbert}.

Unlike supervised learning, which depends on labelled training data, unsupervised WSD typically induces senses directly from corpora, using methods such as clustering of contextual embeddings, graph-based sense induction, and substitute-based approaches that group occurrences by distributional similarity~\citep{martinez2023context}.


Knowledge-based approaches leverage computational lexicons such as WordNet and BabelNet, utilising their graph structure, where synsets serve as nodes and relationships between them act as edges. Successful methods in this category frequently employ graph algorithms like random walks~\citep{agirre2014random}, clique approximation~\citep{moro2014entity}, or game-theoretic models~\citep{tripodi2019game}. The effectiveness of such approaches is dependent mainly on the richness and accuracy of the knowledge base they are built upon~\citep{pilehvar2014large, maru2019syntagnet}.

Among the top-performing models, two distinct approaches stand out: SyntagRank~\citep{scozzafava2020personalized} and SREF\textsubscript{KB}~\citep{wang2020synset}. SyntagRank is a fully graph-based approach that applies the Personalised PageRank algorithm~\citep{page1999pagerank} to an enhanced version of BabelNet's WordNet component, integrating additional relations from the WNG corpus and SyntagNet~\citep{maru2019syntagnet}. This resource captures collocation-based sense relations.

Conversely, SREF\textsubscript{KB} is a vector-based method that utilises contextualised word representations and sense embeddings for disambiguation. It generates sense vectors by applying BERT~\citep{devlin2019bert} to WordNet definitions and example sentences, as well as automatically retrieved contextual data from the Web. While SyntagRank, powered by BabelNet, demonstrates multilingual scalability, SREF\textsubscript{KB} has been evaluated primarily in English. Additionally, SREF\textsubscript{KB} integrates manually curated example sentences from WordNet, introducing a degree of supervised learning, which may contribute to its improved performance. 

It is important to note that in this study, WSD is not treated as a standalone task. 
Instead, it serves as a necessary stage within our pipeline for LSCD, 
enabling the disambiguation of slang word senses before analysing how these senses evolve over time.



\subsection{Methods for Lexical Semantic Change Detection (LSCD)}
\label{subsec:bdlsc_scd}

Lexical Semantic Change Detection (LSCD) is the task of identifying whether the meaning of a word has shifted across different time periods~\citep{tahmasebi2021survey, schlechtweg2020semeval}. 
Formally, given a target word \(w\) and corpora representing two time intervals \(t_1\) and \(t_2\), the goal is to determine changes in the sense distribution of \(w\) between \(t_1\) and \(t_2\)~\citep{hamilton2016cultural, kutuzov2018diachronic, schlechtweg2020semeval, tahmasebi2021survey}.

Building on longstanding linguistic interest in how meanings evolve over time~\citep{traugott2001regularity}, recent advances in diachronic corpora and word representation techniques have enabled automated approaches to LSCD in Natural Language Processing (NLP). Unsupervised LSCD approaches primarily identify semantic shifts by comparing word embeddings trained on corpora from different time periods. Several techniques exist for aligning vector spaces across time, including initialisation-based methods~\citep{kim2014temporal}, alignment techniques~\citep{kulkarni2015statistically, hamilton2016cultural}, and joint learning models~\citep{yao2018dynamic, dubossarsky2019timeout}.

In addition to static word embeddings, researchers have developed methods for analysing contextualised word representations by either averaging multiple word embeddings~\citep{martinc2020leveraging,   laicher2021explaining} or comparing individual embedding pairs~\citep{ laicher2021explaining}. Furthermore, approaches considering variance in embedding distributions have been proposed~\citep{aida2023unsupervised, nagata2023variance}.

These unsupervised LSCD methods are often evaluated using benchmark tasks designed to measure semantic change~\citep{schlechtweg2020semeval}. Additionally, the Cross-Lingual Lexical Semantic Change Model (XL-LEXEME), fine-tuned on Word-in-Context (WiC) datasets, has demonstrated competitive performance on specific LSCD tasks. Recent studies indicate that large language models (LLMs), such as GPT-3.5, still struggle with LSCD, particularly in short-term contexts, even when optimised prompts are applied~\citep{periti2024chat}. However, GPT-4 has improved LSCD performance when using well-structured few-shot learning approaches~\citep{ren2024few}.

The Unsupervised LSCD task~\citep{schlechtweg2020semeval} is a SemEval challenge that compares approaches for detecting semantic change. The task provides a unified framework and standardised dataset from the Clean Corpus of Historical American English (CCOHA)~\citep{alatrash2020ccoha} to enable fair comparisons, as previous studies have used different procedures, languages, and corpora. The task involves determining whether a set of target words has undergone semantic change across two distinct periods. The dataset includes corpora for German, English, Latin, and Swedish, as well as two subtasks: binary classification of target words as having or not having changed their meaning and ranking of target words based on the degree of semantic change.

As the SemEval subtask one described above is most relevant to our work, we reviewed the performance of the best-performing methods officially evaluated by the task organisers. The best outcome for English was an accuracy of 0.730. Their method is underpinned by contextualised word embeddings provided by a multilingual BERT (mBERT) model~\citep{devlin2019bert}. They applied the {Uniform Manifold Approximation and Projection (UMAP) algorithm to reduce the dimensionality of the embeddings, which allowed for a broader range of clustering algorithms that could automatically determine the number of clusters~\citep{rother2020cmce}. The researchers compared the performance of Hierarchical Density-Based Clustering (HDBSCAN) to that of Gaussian Mixture Models (GMMs) and other clustering algorithms. Remarkably, the model performed well on binary classification despite using only 10-dimensional mBERT embeddings (reduced from the original 768 dimensions).

Very few papers have explicitly framed binary LSCD as a word sense
disambiguation problem. Teodorescu et al.~\citep{teodorescu-etal-2022-black}
cast binary change detection as WSD for Spanish in the LSCDiscovery shared
task, implementing the approach using
AMuSE~\footnote{\url{http://nlp.uniroma1.it/amuse-wsd/about}}, a
user-friendly end-to-end neural WSD system that incorporates pre-processing
steps such as tokenisation, lemmatisation, and parts-of-speech tagging. The
system assumes that word semantics have changed if a sense is observed in only
one of the two corpora (label~1) or if the relative change for any sense
exceeds a tuned threshold (set at 0.65) (label~1); otherwise, it concludes
that there is no change (label~0). Tang
et al.~\citep{tang2023can} applied a comparable sense-distribution approach to
English, German, Swedish, and Latin on the SemEval-2020 Task~1 binary subtask,
using pretrained static sense embeddings to disambiguate each occurrence and
comparing sense-frequency distributions across periods. Both studies
demonstrate that WSD-derived sense distributions can serve as a basis for
binary change classification, yet the approach remains underexplored relative
to embedding-based methods.




\subsection{Research Gaps}
\label{subsec:bdlsc_gaps}

Our review of prior work on both WSD and LSCD highlights several significant gaps that motivate our study.

For WSD, existing research has made substantial progress through supervised, unsupervised, and knowledge-based methods. However, evaluations have been mainly based on traditional benchmarks such as SemCor, WordNet, and FEWS. These resources primarily capture standard, stable vocabulary and do not account for words that exhibit both slang and non-slang senses. Such words are often highly polysemous, context-dependent, and rapidly evolving. As a result, it remains unclear how current WSD methods perform when applied to dynamic and non-standard lexical items, leaving a critical gap in the literature.

With respect to LSCD, prior work has predominantly cast the task as binary classification, deciding whether a word has changed between two time periods. Unsupervised embedding-based methods are the most common, but they struggle to capture more nuanced trajectories of change, such as senses that simultaneously emerge and disappear. Supervised techniques are rare, mainly due to the lack of annotated corpora, and there is still no widely adopted benchmark for studying sense change beyond binary settings.

While prior work has examined slang
formation~\citep{kulkarni2018simple,sun2021computational},
neologism detection~\citep{lau2012word,cook2014novel}, and short-term
lexical innovation~\citep{keidar2022slangvolution,stewart2018making},
to our knowledge no existing resource
 provides instance-level, per-period sense annotations for words that carry both slang and standard meanings simultaneously; it is this specific gap that our datasets address. In this work, we directly address these gaps. We present BD-LSC, the first dataset of words annotated across three time periods where each word contains both slang and non-slang senses, with labels for \textit{sense gain}, \textit{sense loss}, and \textit{no change}. We also present the ST-WSD Dataset, which provides instance-level annotations for words with both slang and non-slang senses and supports fine-grained evaluation of WSD and LSCD. We evaluate both supervised and unsupervised approaches to WSD in this mixed-sense context, and we systematically benchmark methods for LSCD, spanning clustering-based techniques, baseline classifiers, and large language models. In doing so, we provide the first benchmark for studying the interaction between slang and non-slang senses and establish a reproducible framework for future research.


\section{Dataset Creation}
\label{sec:bdlsc_dataset_creation}


\subsection{Available Datasets}
\label{subsec:bdlsc_available_datasets}

The Google Books Ngram Corpus~\citep{lin2012syntactic} is an extensive diachronic resource containing N-grams extracted from digitised books spanning several centuries. It enables the analysis of long-term lexical trends and offers subcorpora such as British English, American English, and Fiction. However, it suffers from substantial OCR noise, particularly in older texts, and provides only N-gram snippets rather than complete sentences, which restricts the study of contextual semantic change.

The Corpus of Historical American English (COHA)~\citep{davies2012expanding} is another widely used diachronic corpus, containing 400 million words from 1810 to 2000 across fiction, magazines, newspapers, and non-fiction. A cleaned version, CCOHA~\citep{alatrash2020ccoha}, is commonly used in LSCD research. COHA is organised into decade-level groupings, making it suitable for long-term comparison. Its companion resource, the Corpus of Contemporary American English (COCA)~\citep{davies2010corpus}, covers 1990 to 2017 and provides one billion words across a wide range of genres, offering a more modern perspective.

Additional diachronic resources include the DiAchronic TExt (DATE) corpus, which spans 1700 to 2010 and incorporates material from CLMET 3.0 and COHA as well as data from SemEval-2015 Task 7~\citep{popescu2015semeval}. The CLMET 3.0 contains approximately 34 million words from 1710 to 1920. Other available datasets, such as the Times Digital Archive and the New York Times Annotated Corpus, offer extensive newspaper material but are limited in genre diversity and are not designed for studying detailed semantic change.
The SemEval-2020 Task 1 dataset~\citep{schlechtweg2020semeval} is currently the most widely used benchmark for lexical semantic change detection. It contains English, German, Latin, and Swedish data drawn primarily from CCOHA. Words are annotated for binary change and graded change ranking between the nineteenth and twentieth centuries, and the dataset remains the standard reference point for evaluating LSCD systems.

More recently, two Word-in-Context (WiC) style datasets have been introduced for diachronic evaluation: TempoWiC and HistoWiC. TempoWiC~\citep{loureiro2022tempowic} focuses on short-term meaning change in social media. It frames the task as a binary WiC problem on paired tweets sampled around frequency peaks. HistoWiC~\citep{periti2024chat} repurposes the English SemEval-2020 data in a WiC format, pairing sentences from distinct historical periods to assess long-term semantic drift. These datasets complement earlier corpora by providing contextualised, sentence-level annotations.

Despite the availability of numerous resources, existing datasets have limitations that restrict their usefulness for studying words that exhibit both slang and non-slang sense usages. Most, including the SemEval-2020 Task 1 dataset, frame LSCD as a binary decision and cannot capture cases in which senses emerge and disappear simultaneously. Many LSCD datasets rely on historical literary corpora such as COHA or the Google Books Ngram Corpus, which do not reflect the rapid evolution of slang in online and conversational contexts. Frequency-based approaches without contextual sense annotation make subtle meaning shifts difficult to detect, and resources such as DWUGs~\citep{schlechtweg2021dwug}, although manually annotated, cover only a small set of lexical items. Furthermore, most corpora track change over decades or centuries, whereas slang often shifts meaning within only a few years. Finally, reliance on predefined sense inventories, such as WordNet, limits coverage of emerging slang senses, many of which are absent from traditional lexicons.



\subsection{Target Word Selection Methodology}
\label{subsec:bdlsc_target_words}

\begin{figure*}[!hbpt]
    \resizebox{\textwidth}{!}
  {%
\includegraphics[width=\textwidth]{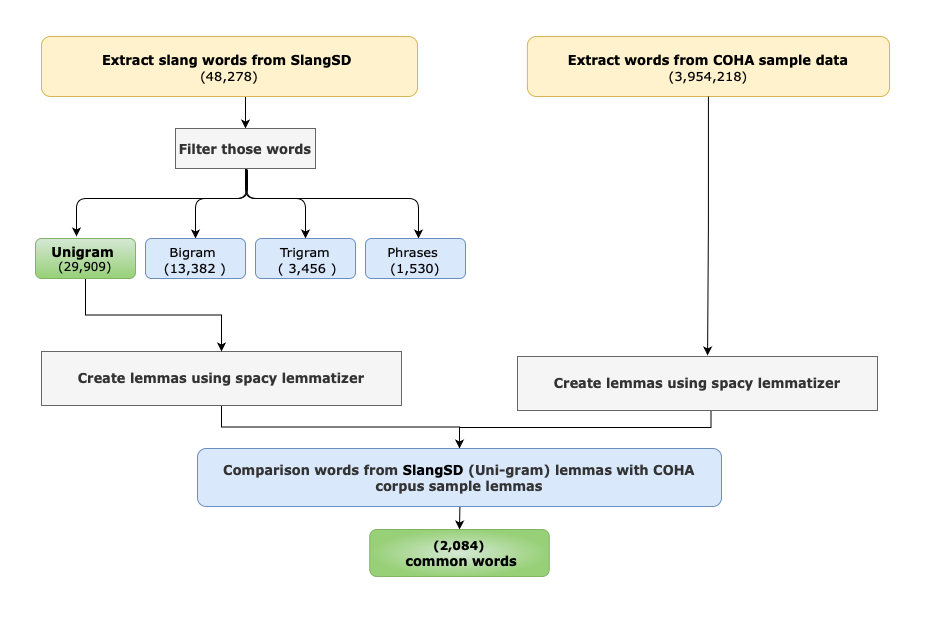}
 }
\caption{Procedure for selecting target words from SlangSD and COHA.}
\label{Fig:1.png}
\end{figure*}

We created a collection of target words (words that exhibit both slang and non-slang sense usages) and may have undergone changes in meaning over time. Specifically, they may have lost or gained a word sense, or their senses may have remained unchanged between the periods T1 and T2, T2 and T3, and T1 and T3.

In our study, the selection process for target slang words was meticulously structured. We initially compiled a comprehensive list from SlangSD~\footnote{\url{https://www.rdocumentation.org/packages/lexicon/versions/1.2.1}}, comprising 48,278 slang instances. This list was methodically categorised into N-gram formations: 29,909 unigrams, 13,382 bi-grams, 3,456 tri-grams, and 1,530 slang phrases. For this study, we focused exclusively on unigrams which were subsequently lemmatised to facilitate direct comparison with the COHA dataset, containing 169,077 lemmas.

To ensure chronological consistency, we compared these lemmatised slang unigrams against the COHA dataset, identifying 2,084 words common to both SlangSD and COHA. Duplicates from this common list were removed to maintain data integrity. The next phase involved a meticulous manual review, cross-referencing words against multiple COHA periods to ensure that only those appearing across several periods were considered. Terms unique to a single period were excluded, as they would not allow for comparative analysis, as illustrated in Figure~\ref{Fig:1.png}.

\begin{figure*}[t]
    \resizebox{\textwidth}{!}
  {%
\includegraphics[width=\textwidth]{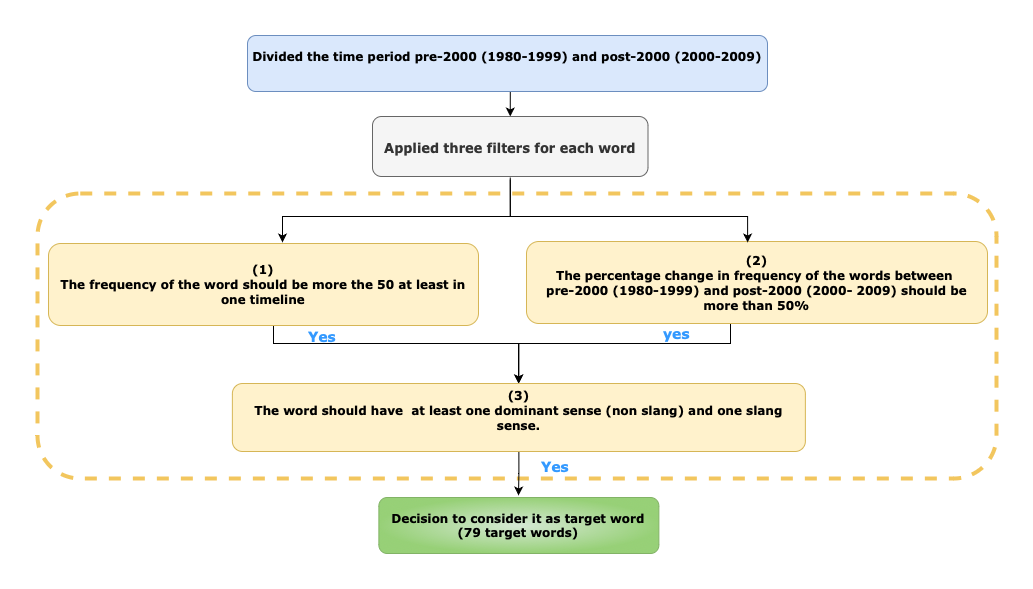}
}
\caption{Process for filtering target words to find suitable candidates for the Lexical semantic change detection task.}
\label{Fig:2.png}
\end{figure*}

To refine our list, we first applied quantitative criteria designed to identify terms that exhibited both temporal persistence and meaningful variation in usage. Words were required to meet the following conditions: 
(1) They appeared with a minimum frequency of 50 in at least one examined period, ensuring sufficient contextual diversity for reliable sense annotation. The 50-occurrence minimum is a threshold we set for this study to ensure each target word carries enough contextual evidence to distinguish stable, emerging, and disappearing senses. It reflects the general principle, common in lexical semantic change research, that very low-frequency items provide insufficient evidence for reliable cross-period comparison~\citep{kutuzov2018diachronic, hamilton2016diachronic}; Hamilton et al., for instance, discarded words occurring fewer than 500 times per yearly slice in the Google Ngram data (100 in COHA) when constructing their diachronic embeddings. The SemEval-2020 Task~1 gold-standard labels likewise apply frequency thresholds ($k=2$, $n=5$ for English, German, and Swedish) to prevent small random fluctuations in sense frequencies from being misclassified as change~\citep{schlechtweg2020semeval}. The specific value of 50 is calibrated to our corpora, which are substantially larger than the annotation samples used in SemEval-2020 Task~1.
(2) They demonstrated a significant variation in frequency defined as a change of at least 50\% between periods, as smaller fluctuations typically reflect corpus-level noise or genre imbalance rather than genuine linguistic trends, whereas a shift of this magnitude offers more substantial evidence of meaningful changes in usage over time.

After a word met these frequency-based criteria, we conducted a qualitative review to ascertain its relevance to slang. This step involved verifying that (3) the word was employed within a slang and non-slang senses. To ensure the integrity of this selection, words lacking a recognised slang sense were excluded. 
By adhering to this sequential filtering process, we ensured that the final selection of words satisfied the statistical thresholds for robust longitudinal analysis and reflected the evolving nature of slang in widespread use. The selection process of the keywords is illustrated in Figure~\ref{Fig:2.png}. After applying these quantitative and qualitative filters, the final BD-LSC set comprised 79 target words, each attested across multiple periods and carrying at least one slang and one non-slang sense.


\subsection{Sense Inventory Construction}
\label{subsec:bdlsc_sense_inventory}

A sense inventory is a list of potential meanings for a given word. The most well-known sense inventories for the English language include: Princeton WordNet~\citep{miller1990introduction} is a comprehensive, manually curated database of English that serves as the de facto standard for WSD. WordNet is structured as a graph, with nodes representing synsets, which are groups of synonymous words in specific contexts. Each word in a synset signifies a distinct sense. Synsets are interconnected through lexical-semantic relations, such as hypernymy (is-a) and meronymy (part-of). Additionally, WordNet provides lexical resources like definitions (glosses) and usage examples. Most recent English WSD studies use version 3.0 (released in 2006), which includes 117,659 synsets. The English WordNet 2020~\citep{mccrae2020english} further expanded the original WordNet by adding approximately 3,000 new synsets, including slang and neologisms.

BabelNet~\citep{navigli2012babelnet} is a multilingual dictionary that includes both lexicographic and encyclopaedic terms, created by semi-automatically integrating various resources, such as WordNet, multilingual versions of WordNet, and Wikipedia. BabelNet is organised as a semantic network, with nodes representing multilingual synsets (groups of synonyms in different languages) and edges denoting semantic relations between them. The latest release, version 5.0 (2021), covers 500 languages and contains over 20 million synsets~\citep{navigli2012babelnet}.

However, these sense inventories are unsuitable for our purposes, as they do not contain all the slang senses for our target words. Therefore, we created our own sense inventory, collecting standard senses for each word from the Oxford English Dictionary\footnote{\url{https://www.oed.com}} and slang senses from Green's Dictionary of Slang~\footnote{\url{https://greensdictofslang.com}}, Urban Dictionary~\footnote{\url{https://www.urbandictionary.com}}, and the Online Slang Dictionary~\footnote{\url{http://onlineslangdictionary.com}}.

A notable critique of Urban Dictionary is its reliance on 
crowdsourced content, which can sometimes compromise the 
reliability of its definitions. To address this, we supplemented 
Urban Dictionary with GDoS and the Online Slang Dictionary, and 
retained a crowdsourced sense only if it was independently attested 
in another source and supported by at least five corpus 
occurrences. More generally, across all sources, closely related 
senses were merged or kept distinct according to their semantic 
function, register, and domain. The full construction and 
consolidation process, including the treatment of proper-noun and 
collocational uses, is detailed in 
Appendix~\ref{app:bdlsc-sense-inventory}.


\subsection{The Bi-Directional\protect\footnote{We use the term \emph{bi-directional Lexical semantic change} to refer to cases where words may simultaneously experience the emergence of new senses and the disappearance of existing senses across time periods.} LSC (BD-LSC) Dataset}
\label{subsec:bdlsc_dataset}

A thorough search was conducted to find a suitable dataset for our task; however, we found no existing English corpora that support detecting semantic change in words that exhibit both Slang and non-slang usage. 

We therefore developed our own corpus by gathering data from multiple sources. The study follows the same methodological steps as outlined in~\citep{schlechtweg2020semeval}, and the dataset includes all instances of contextual information related to the target words. Specifically, we collected documents from three time periods. The first two periods were sourced from the Corpus of Historical American English (COHA)~\citep{davies2012expanding}. The first period, \texttt{T1}, covers the years 1980--1999, while the second period, \texttt{T2}, spans 2000--2009. The third period, \texttt{T3}, consists of Twitter data covering the years 2010--2020.

Our use of three time periods spanning historical and contemporary
registers is consistent with longstanding methodological findings that
multi-point time-series evaluation is preferable to two-point comparison
\cite{shoemark2019room}, and with recent critiques of two-period
benchmark design~\citep{phan2026evaluating}.

The selection of these temporal periods (T1: 1980--1999, T2: 2000--2009, T3: 2010--2020) was guided by linguistic and corpus-driven considerations rather than arbitrary boundaries. T1 corresponds to the pre-Internet era, during which slang is relatively scarce in historical written corpora such as COHA. Preliminary frequency checks showed that slang-like usages appear 2--5 times less often in COHA before 2000 than in post-2000 COHA or Twitter data, reflecting the limited representation of informal registers in this earlier period. Although COHA’s 1980s and 1990s slices each contain substantial token counts (approximately 25-28 million words per decade), slang remains markedly under-represented. Combining these two decades into a single 20-year interval, therefore, increases the number of available contexts while preserving linguistic continuity, ensuring sufficient evidence for reliable sense comparison and annotation.

The period \texttt{T2} (2000--2009) was intentionally defined as a shorter 10-year interval to isolate early internet-era usage before the widespread adoption of social media. Finally, \texttt{T3} (2010--2020), drawn from Twitter, captures the evolution of contemporary slang driven by online platforms, where new meanings emerge rapidly. This temporal structure provides a balanced, theoretically grounded framework for analysing semantic change across the pre-internet, early-internet, and social-media eras.

The clean version of CCOHA~\citep{alatrash2020ccoha} was chosen for periods \texttt{T1} and \texttt{T2} due to its high-quality preprocessing and extensive documentation. COHA’s decade-level organisation offers an advantageous chronological framework that enables precise comparative analyses. The inclusion of Twitter for \texttt{T3} ensures that the dataset captures modern, dynamic slang usage characteristic of digital communication platforms.

The resulting BD-LSC dataset is publicly available~\footnote{\url{https://github.com/Afnan-Aloraini/Bi-Directional-Lexical-Semantic-Change-Dataset}} and serves as a resource for researchers interested in lexical semantic change, slang evolution, and sense-level semantic analysis across distinct stages of recent linguistic history.


\subsection{Annotation Guidelines for BD-LSC}
\label{subsec:bdlsc_annotation}

For the full dataset, annotators were tasked with examining words and their senses within each specified period (T1, T2, and T3, as defined earlier). The detailed annotation procedure included:

\begin{itemize}
    \item Annotators were provided with all instances of each target word appearing in the COHA and Twitter corpora, spanning three distinct periods (T1--T3).
    \item Each annotator had access to the complete list of existing word senses, encompassing both Slang and non-slang interpretations.
    \item The annotators’ task was to determine the \emph{presence} or \emph{absence} of each word sense within the provided instances for the specific time period.
    \item A sense was considered \emph{present} in a given time period if it appeared in at least five unambiguous instances of usage within that period.
\end{itemize}

All instances of each target word across the whole corpus (T1--T3) were independently labelled by two annotators, with a third serving as primary annotator and adjudicator, consistent with the procedure described in Section~\ref{subsec:inter_annotator_agreement}. The results of the annotation process, as presented by the annotator, detail the incidence of each word's sense throughout different time periods in our dataset. This is systematically depicted in a format analogous to Table~\ref{tab:sense_timelines}, employing a binary system for clarity: a~\say{1} indicates that a specific sense of a word has been detected within a time period. In contrast, a ~\say{0} shows that the sense was absent during that period. This method provides a clear, concise record of word-sense occurrences across the dataset.


\subsection{Annotation Example}
\label{subsec:bdlsc_example}

The target word is \textit{mammy}, which has four identified senses mentioned below:
\begin{itemize}
    \item Sense 1: A stereotypical depiction of a Black woman employed as a nanny or cook in white households.

    \item Sense 2: A colloquial or dialectal term meaning ‘mother’.

    \item Sense 3: Slang meaning ‘a large amount’ (e.g., money’s mammy = ‘a great deal of money’).
    
    \item Sense 4: A proper noun referring to a brand, company, or artistic title.

\end{itemize}

Table~\ref{tab:occurrences_target_word} showcases our evaluation results, tracking the occurrences of the word \textit{mammy} and its senses across three distinct periods in our dataset. We used a binary system for documentation: ~\say{1} indicates that a sense was observed, and ~\say{0} demonstrates that it was not found. The results show that Sense 1 and Sense 2 of \textit{mammy} were present in all three time periods. Sense 4 was present in T1, absent in T2, and present again in T3, whereas Sense 3 appeared only in T3.

\begin{table}[!h]
\centering
\captionsetup{width=\textwidth} 
\caption{Occurrences of target word by time period. {Note.} S1--S4 refer to distinct dictionary senses of the target word.}
\renewcommand{\arraystretch}{1.2}
\setlength{\tabcolsep}{12pt}
\begin{tabular}{c cccc}
    \hline
    \multicolumn{5}{c}{\small Target Word: Mammy} \\
    \hline
    {~\footnotesize Period} & {~\footnotesize S1} & {~\footnotesize S2} & {~\footnotesize S3} & {~\footnotesize S4}\\
    \hline
    {~\footnotesize 1980-1999}  & 1 & 1 & 0 & 1\\
    {~\footnotesize 2000-2009}  & 1 & 1 & 0 & 0\\
    {~\footnotesize 2010-2020}  & 1 & 1 & 1 & 1\\
    \hline
\end{tabular}
\label{tab:occurrences_target_word}
\end{table}


\subsection{SlangTrack Word Sense Disambiguation (ST-WSD) Dataset}
\label{subsec:stwds_dataset}

For our experiments, we selected a subset of ten words from the full dataset, following criteria designed to ensure both analytical tractability and semantic diversity. Each word contains eight or fewer senses, and together the set reflects a broad range of semantic change behaviours, including stability, sense addition, sense loss, and cases exhibiting both types of change.

The relatively small number of target words was a deliberate methodological choice. Each occurrence required full manual sense annotation across three time periods, resulting in more than 12,000 labelled instances. COHA contexts were extracted at the paragraph level, often including several sentences before and after the target sentence. This expanded window was necessary because, in COHA, sense-disambiguating cues particularly those associated with slang or figurative usage frequently appear multiple sentences away from the target word. Consequently, each instance consisted of a multi-sentence passage requiring careful and time-intensive reading, which made the annotation workload substantially higher than in standard WSD datasets based on isolated sentences. A ten-word subset therefore provided a realistic balance between annotation quality, semantic coverage, and practical feasibility.

Table~\ref{tab:word_frequencies} reports statistics for the selected words, including sense inventories, instance counts per period, and semantic change labels across T1--T2, T2--T3, and T1--T3. We note that genuinely simultaneous gain and loss within a single period pair is comparatively rare in this ten-word subset (observed for two words, both at T1--T2); most change is single-directional. The bi-directional formulation is nonetheless necessary to represent these cases without forcing them into a binary changed/stable decision, and the full BD-LSC set contains further such instances (for example, Battery, Cheese, Sketch, and Skinny in Appendix~\ref{app:bdlsc-corpus-overview}).Table~\ref{tab:sense_timelines} lists all senses and their attested presence across time periods. The dataset is publicly available~\footnote{\url{https://github.com/SlangTrack/SlangTrack-Word-Sense-Disambiguation}} as a resource for research on LSCD, slang, and WSD.

\begin{table}[!h]
\centering
\captionsetup{width=\textwidth} 
\caption{Word frequency statistics for ST-WSD  Dataset. Key: NC = no change, SA = sense added, SR = sense removed.}
\renewcommand{\arraystretch}{1.2}
\setlength{\tabcolsep}{4pt}
\small
\begin{tabular}{c c ccc ccc}
    \toprule
    Word & \makecell{No. of \\ Senses} & 
    \makecell{Sentences \\ T1} & 
    \makecell{Sentences \\ T2} & 
    \makecell{Sentences \\ T3} & 
    \makecell{Labels \\ T1-T2} & 
    \makecell{Labels \\ T1-T3} & 
    \makecell{Labels \\ T2-T3} \\ 
    \midrule
    Eat      & 6 & 975  & 995  & 1000 & NC     & NC       & NC \\ 
    BMW      & 3 & 91   & 141  & 865  & SR     & SA       & SA \\ 
    Brownie  & 8 & 73   & 72   & 819 & SA     & SA       & SA \\ 
    Chronic  & 5 & 482  & 380  & 823  & SA     & SA       & SA \\ 
    Climber  & 4 & 75   & 50   & 517  & NC     & SA       & SA \\ 
    Germ     & 6 & 112  & 70   & 650  & SA, SR & SA       & SA \\ 
    Mammy    & 4 & 172  & 22   & 876  & SR      & SA       & SA \\ 
    Cucumber & 3 & 74   & 85   & 892  & SA, SR & SA       & SA \\ 
    Rodent   & 2 & 101  & 67   & 905  & NC     & NC       & NC \\ 
    Salty    & 6 & 214  & 165  & 887  & NC     & NC       & NC \\ 

    \bottomrule
\end{tabular}
\label{tab:word_frequencies}
\end{table}


\begin{table*}[!hbtp]
\centering
\caption{ Target words from the ST-WSD dataset, their dictionary senses, and their occurrences across three time periods (T1, T2, T3).}
\renewcommand{\arraystretch}{1.4}
\resizebox{\textwidth}{!}{
\begin{tabular}{c p{12cm} c c c}
\hline
\textbf{Word} & \multicolumn{1}{c}{\textbf{Senses from Dictionary}} & \textbf{T1} & \textbf{T2} & \textbf{T3} \\
\hline
\textbf{Eat} &
1. To consume food or any substance.
2. To perform oral sex on a woman.
3. To make money or to absorb a financial loss (“take the hit”).
4. To rob someone in the street in a low-violent manner.
5. To defeat, overwhelm, or destroy someone or something.
6. To irritate, bother, or annoy. &
\makecell{S1, S2, S3 \\ S4, S5, S6} &
\makecell{S1, S2, S3 \\ S4, S5, S6} &
\makecell{S1, S2, S3 \\ S4, S5, S6} \\
\hline
\textbf{BMW} &
1. A German luxury automobile brand.
2. A derogatory reference to Black individuals (e.g., “Black Man Working,” “Black Magic Woman”).
3. An acronym for “Be My Wife,” used as a suggestive or intimate invitation. &
\makecell{S1, S2} &
S1 &
\makecell{S1, S2 \\ S3} \\
\hline
\textbf{Brownie} &
1. A chocolate-baked dessert.
2. A person who behaves foolishly.
3. A racial insult referring to individuals with brown skin.
4. A marijuana-infused edible.
5. A vulgar reference to the anus.
6. A symbolic or imaginary unit of social credit (“brownie points”).
7. A personal name or nickname.
8. A frequent collocate used in set expressions. &
\makecell{S1, S2, S3 \\ S7, S8} &
\makecell{S1, S2, S3 \\ S4, S7, S8} &
\makecell{S1, S2, S3 \\ S4, S5, S6, \\S7, S8} \\
\hline
\textbf{Chronic} &
1- Very high-quality weed, a reference to a drug. 2- Pertaining to a long-lasting
medical condition. 3- Excellent. 4- A person having a particularly bad habit,
extreme, used in a negative sense and often as something chronic. 5- Name of
something, i.e. album, song, etc. &
\makecell{S1, S2, S4} &
\makecell{S1, S2, S3 \\ S4} &
\makecell{S1, S2, S3 \\ S4, S5} \\
\hline
\textbf{Climber} &
1. A person who seeks to rise socially by associating with influential individuals.
2. A person who climbs as a sport (rock climber).
3. A plant that grows upward by attaching to supports.
4. A burglar who enters buildings by climbing, especially to access higher floors. &
\makecell{S1, S2, S3} &
\makecell{S1, S2, S3} &
\makecell{S1, S2, S3 \\ S4} \\
\hline
\textbf{Germ} &
1. A microorganism capable of causing disease.
2. A term referring to a cigarette within certain institutional settings.
3. A derogatory term used to refer to individuals of German background.
4. A term for a contemptible or unpleasant person.
5. A proper noun referring to companies, brands, or trademarks.
6. A collocate commonly used as part of compound expressions.&
\makecell{S1, S3, S4 \\ S6} &
\makecell{S1, S3, S5 \\ S6} &
\makecell{S1, S2, S3 \\ S4, S5, S6} \\
\hline
\textbf{Mammy} &
1. A stereotypical depiction of a Black woman employed as a nanny or cook in white households.
2. A colloquial or dialectal term meaning ‘mother’.
3. Slang meaning ‘a large amount’ (e.g., money’s mammy = ‘a great deal of money’).
4. A proper noun referring to a brand, company, or artistic title. &
\makecell{S1, S2, S4} &
\makecell{S1, S2} &
\makecell{S1, S2, S3 \\ S4} \\
\hline
\textbf{Cucumber} &
1. A term referring to the penis.
2. A long green vegetable with watery flesh and edible skin.
3. A proper noun referring to companies, brands, or software names. &
\makecell{S1, S2} &
\makecell{S2, S3} &
\makecell{S1, S2, S3} \\
\hline
\textbf{Rodent} &
1. An insult for someone considered unattractive, untrustworthy, or unintelligent.
2. A mammal of the order Rodentia, including rats, mice, hamsters, and squirrels. &
\makecell{S1, S2} &
\makecell{S1, S2} &
\makecell{S1, S2} \\
\hline
\textbf{Salty} &
1. Irritated, annoyed, resentful, or “sour.” 
2. Containing salt or having a salty flavour.
3. Old, worn, or well-used, sometimes with a positive connotation.
4. Tough, aggressive, or hardened (e.g., a seasoned or experienced person).
5. Crude, obscene, or vulgar.
6. A proper noun referring to names of animals, brands, songs, or similar entities. &
\makecell{S1, S2, S3 \\ S4, S5, S6} &
\makecell{S1, S2, S3 \\ S4, S5, S6} &
\makecell{S1, S2, S3 \\ S4, S5, S6} \\
\hline
\end{tabular}
}
\label{tab:sense_timelines}
\end{table*}



\subsubsection{Annotation of SlangTrack Word Sense Disambiguation (ST-WSD) Dataset}
\label{subsubsec:stwsd_annotation}

For the ST-WSD dataset, we undertook a meticulous annotation of each instance. Annotators were tasked with assigning the most appropriate sense to every occurrence of the target words, following a standard WSD-style procedure. Unlike approaches that simply identify the presence or absence of particular senses within a time period, our annotation required evaluating each instance individually to determine its specific sense. This ensured a nuanced and precise representation of semantic behaviour across the dataset.



\subsection{Annotation Procedure and Inter-Annotator Agreement for BD-LSC and ST-WSD Datasets}
\label{subsec:inter_annotator_agreement}

Our annotation team comprised three individuals, each with strong English proficiency and a bachelor's degree or higher. One held a degree in Linguistics and served as the primary annotator and adjudicator. A pilot annotation task involved two annotators independently labelling a random sample of 100 sentences from each time period; this pilot phase served as the basis for refining the annotation guidelines to ensure consistent and reliable results. Following the established guidelines in Section~\ref{subsec:bdlsc_annotation}, two annotators then independently labelled the full corpus across all periods, and the primary annotator subsequently adjudicated their disagreements to produce the final gold labels. Inter-annotator agreement was computed on the two independent label sets, before adjudication, using Cohen's $\kappa$.

For the BD-LSC dataset, agreement was 0.92 for T1, 0.89 for T2, and 0.86 for T3, indicating an almost perfect level of agreement among annotators.
For the ST-WSD dataset, the corresponding values were 0.90, 0.89, and 0.87 for T1, T2, and T3, respectively.


\subsection{Comparative Overview of LSC Datasets}
\label{subsec:bdlsc_comparison}

To contextualise the datasets introduced in this study within the broader field of lexical semantic change (LSC) research, 
Table~\ref{tab:lsc_combined_part1} presents an overview of major LSC benchmarks. 
This table includes established datasets such as ~SemEval 2020, TempoWiC, 
HistoWiC, and DWUG, alongside the new BD-LSC and~SlangTrack-WSD datasets introduced in this work.

The comparison highlights key distinctions in temporal coverage, supervision, sense inventory, and annotation methodology. 
Traditional datasets, such as ~SemEval 2020 and DWUG, focus on long-term historical corpora and graded LSCD. 
WiC-style benchmarks, such as TempoWiC and HistoWiC, enable contextualised binary meaning comparison over time. 
By contrast, the BD-LSC and SlangTrack-WSD datasets uniquely capture slang-driven and bi-directional semantic change, incorporating explicit sense annotation and WSD across three temporal periods. 
These properties extend the scope of LSC evaluation to include informal and contemporary language change.

For clarity, only datasets explicitly designed for semantic change evaluation are included in the comparison. 
We excluded datasets such as Google Ngram, COHA, COCA, DATE, and CLMET 3.0 as these are general-purpose or diachronic corpora that serve as data sources but do not include semantic change annotations or evaluation protocols.




\begin{sidewaystable*}[p]
\centering
\caption{Comparison of major lexical semantic change (LSC) datasets.}
\scriptsize
\setlength{\tabcolsep}{1pt}
\renewcommand{\arraystretch}{2.15}

\begin{tabularx}{\textheight}{@{}p{2.2cm} p{2.6cm} *{6}{>{\centering\arraybackslash}X}@{}}
\toprule
& & \multicolumn{4}{c}{\textbf{Existing Benchmarks}} & \multicolumn{2}{c}{\textbf{Ours}} \\
\cmidrule(lr){3-6}\cmidrule(lr){7-8}
\textbf{Category} &
\textbf{Feature} &
\textbf{SemEval 2020} &
\textbf{TempoWiC} &
\textbf{HistoWiC} &
\textbf{DWUG} &
\textbf{BD-LSC} &
\textbf{SlangTrack-WSD} \\
\midrule

\textbf{Dataset Metadata} &
\textbf{Languages\footnote{Language codes: EN = English, DE = German, LA = Latin, SV = Swedish.}} &
EN, DE, LA, SV &
EN &
EN &
EN, DE, LA, SV &
EN &
EN \\

& \textbf{Temporal Span} &
2 periods (19\textsuperscript{th}--20\textsuperscript{th} c.) &
2019--2021 (short term) &
2 periods (19\textsuperscript{th}--20\textsuperscript{th} c.) &
Multi-epoch\footnote{Multi-epoch = multiple temporal slices ($>2$).} &
3 periods 1980--2020 &
3 periods 1980--2020 \\

& \textbf{Sources\footnote{Non-English corpora used in SemEval 2020 and DWUG are drawn from historical and national resources:
German (DE) -- Deutsches Textarchiv (DTA), Berliner Zeitung, Neues Deutschland;
Latin (LA) -- LatinISE;
Swedish (SV) -- Kubhist.}} &
CCOHA &
Twitter &
CCOHA &
CCOHA &
CCOHA + Twitter &
CCOHA + Twitter \\

& \textbf{Cross-domain Composition} &
\xmark & \xmark & \xmark & \xmark & \cmark & \cmark \\
\midrule

\textbf{Annotation \& Supervision} &
\textbf{Supervision} &
Human (U) &
Human (S) &
Human (U) &
Human (semantic proximity) &
Human + manual sense tracking (S) &
Human + manual sense tracking + WSD (S) \\

& \textbf{Sense Inventory} &
\xmark & \xmark & \xmark & \xmark & \cmark & \cmark \\
\bottomrule
\end{tabularx}

\label{tab:lsc_combined_part1}
\end{sidewaystable*}



\begin{sidewaystable*}[p]
\centering
\caption*{Table~\ref{tab:lsc_combined_part1} (Continued). Comparison of Major Lexical Semantic Change (LSC) Datasets.}

\addtocounter{footnote}{1}
\footnotetext{Abbreviations: U = unsupervised, S = supervised, WiC = Word-in-Context, DWUG = DURel Word Usage Graph.}

\scriptsize
\setlength{\tabcolsep}{3pt}
\renewcommand{\arraystretch}{3.0}

\begin{tabularx}{\textheight}{@{}p{1.7cm} p{2.3cm} *{6}{>{\centering\arraybackslash}X}@{}}
\toprule
\textbf{Category} &
\textbf{Feature} &
\textbf{SemEval 2020} &
\textbf{TempoWiC} &
\textbf{HistoWiC} &
\textbf{DWUG} &
\textbf{BD-LSC (Ours)} &
\textbf{SlangTrack-WSD (Ours)} \\
\midrule

& \textbf{Inter-Annotator Agreement} &
Spearman $\rho$ = 0.74 (EN)\footnote{Full IAA: DE = 0.64, LA = 0.80, SV = 0.68.} &
Fleiss' $\kappa$ = 0.45 (EN)\footnote{TempoWiC reports Fleiss' $\kappa$ = 0.446, Krippendorff’s $\alpha$ = 0.439, max pairwise $\alpha$ = 0.627. Words with $\alpha < 0.1$ removed.} &
Spearman $\rho$ = 0.74 (EN)\footnote{HistoWiC reuses gold annotations from SemEval 2020 and TempoWiC; see respective dataset statistics.} &
Spearman $\rho$ = 0.69 (EN)\footnote{DWUG per-language IAA: DE $\rho$ = 0.59, $\alpha$ = 0.53; SV $\rho$ = 0.57, $\alpha$ = 0.56; LA $\rho$ = 0.64, $\alpha$ = 0.62 (mostly single annotator).} &
Cohen’s $\kappa$ = 0.92--0.86\footnote{BD-LSC: per-period agreement --- T1 = 0.916, T2 = 0.887, T3 = 0.858.} &
Cohen’s $\kappa$ = 0.90--0.87\footnote{SlangTrack-WSD: per-period agreement --- T1 = 0.900, T2 = 0.891, T3 = 0.870.} \\
\midrule

\textbf{Task Design} &
\textbf{Task Setup} &
Binary + graded ranking (LSC) &
Binary (short-term change) &
Binary + ranking (WiC-style) &
Unsupervised clustering (sense induction) &
Multilabel shift (Gain/Loss/Stable) &
WSD + multilabel shift (Gain/Loss/Stable) \\

& \textbf{Bi-directional Change Modelling} &
\xmark & \xmark & \xmark & \xmark & \cmark & \cmark \\

& \textbf{WSD Integration} &
\xmark & \xmark & \xmark & \xmark & \xmark & \cmark \\
\midrule

\end{tabularx}

\label{tab:lsc_combined_part2}
\end{sidewaystable*}


\begin{sidewaystable*}[p]
\centering
\caption*{Table~\ref{tab:lsc_combined_part1} (Continued). Comparison of Major Lexical Semantic Change (LSC) Datasets.}


\scriptsize
\setlength{\tabcolsep}{3pt}
\renewcommand{\arraystretch}{3.0}

\begin{tabularx}{\textheight}{@{}p{1.7cm} p{2.3cm} *{6}{>{\centering\arraybackslash}X}@{}}
\toprule
\textbf{Category} &
\textbf{Feature} &
\textbf{SemEval 2020} &
\textbf{TempoWiC} &
\textbf{HistoWiC} &
\textbf{DWUG} &
\textbf{BD-LSC (Ours)} &
\textbf{SlangTrack-WSD (Ours)} \\
\midrule

\textbf{Data Composition} &
\textbf{Target Words} &
31–48 / language &
33 &
37 &
$\approx$ 168 total &
79 &
10 \\

& \textbf{Availability} &
\cmark & \cmark & \cmark & \cmark & \cmark & \cmark \\
\midrule
\textbf{Semantic Capabilities} &
\textbf{Slang Coverage} &
\xmark & \xmark & \xmark & \xmark & \cmark & \cmark \\

& \textbf{Semantic Traceability (Word \& Sense)} &
\xmark & \xmark & \xmark & \xmark & \cmark & \cmark \\
\bottomrule
\end{tabularx}

\label{tab:lsc_combined_part3}
\end{sidewaystable*}


\section{Methodology}
\label{sec:bdlsc_methodology}

To identify semantic change across different time periods, we employed a range of methodological families: unsupervised clustering based on contextual embeddings, supervised machine learning, fine-tuned transformer models, and large language model prompting methods. This section outlines the pre-processing steps, evaluation metrics, and experimental design.



\subsection{Pre-processing}
\label{subsec:bdlsc_preprocessing}

Our approach followed standard pre-processing procedures, including removing duplicate instances, punctuation, URLs, and usernames, and converting all text to lowercase. We did not apply additional steps such as stopword removal or lemmatisation, as language models perform best when sentences remain in their original form. During the initial filtering stage of text extraction, we also removed all instances where the target word appeared as part of a URL or username.

\subsection{Evaluation}
\label{subsec:bdlsc_evaluation}

Our study used \textit{\textbf{multi-label accuracy (MA)}} scores to evaluate the effectiveness of detecting semantic changes between two time periods. Our evaluation metrics compare the available senses in each time period against the predicted senses in each time period for every word to assign final labels: \textit{\say{Sense Added}}, \textit{\say{Sense Removed}}, and \textit{\say{No Change}} between the two time periods.

Additionally, \textit{\textbf{macro and micro F1 scores}}, computed in the standard way from per-sense precision and recall, were used to assess WSD accuracy across the dataset, with macro-F1 weighting each sense equally and therefore reflecting performance on rare slang senses.

\textit{\textbf{Exact Sense Match (ESM):}} This metric represents the proportion of true senses in a given time period that are correctly predicted by the model. Let \(T_X\) denote either time period \(T_1\), \(T_2\), or \(T_3\).

\begin{equation}
    \text{Exact Sense Match (ESM)} =
    \frac{\text{Number of correctly predicted senses in } T_X}
    {\text{Total number of true senses in } T_X} \times 100
\end{equation}

Low ESM with high MA indicates that the model correctly identifies semantic change, but with an inaccurate set of senses. Conversely, high ESM but low MA suggests that the model accurately identifies senses within a time period but fails to capture a critical sense necessary for change detection. High ESM and high MA together reflect accurate coverage and identification of all pertinent senses, leading to robust MA.

Concretely, the two metrics operate at different levels. ESM is a per-period coverage score: within a single period it measures the proportion of gold senses the model correctly recovers, and it therefore carries no notion of change or direction on its own. MA, by contrast, is derived across a period pair: the presence or absence of every individual sense is compared between the two periods, and because each sense is judged independently, a single word can be labelled with sense gain and sense loss simultaneously (for example, the \say{SA, SR} cases in Table~\ref{tab:word_frequencies}). The binary subtask of SemEval-2020 Task~1~\citep{schlechtweg2020semeval}, although also defined in terms of senses being gained or lost, collapses this into a single change label per word, and so records neither which senses changed nor whether gain and loss co-occur; the graded subtask similarly reduces each word to one degree-of-change score. MA retains this per-sense, directional information, while ESM complements it by quantifying within-period sense coverage, which together is what our bi-directional setting requires.



\subsection{Data Generation and Augmentation (Training Data)}
\label{subsec:bdlsc_data_generation}
 
The training data for our experiments needs to be different from the test data, as our models are expected to perform in real life on unseen data from different modalities and platforms. Our curated WSD dataset, explained in Section~\ref{subsec:stwds_dataset} with three time periods, was used as a test dataset in all our experiments for consistency. No dataset from the test data was added or used in any way in the training data. All training data was synthetically created and augmented. 


 \begin{figure*}[!hpt]
    \resizebox{\textwidth}{!}
{%
\includegraphics[width=\textwidth]{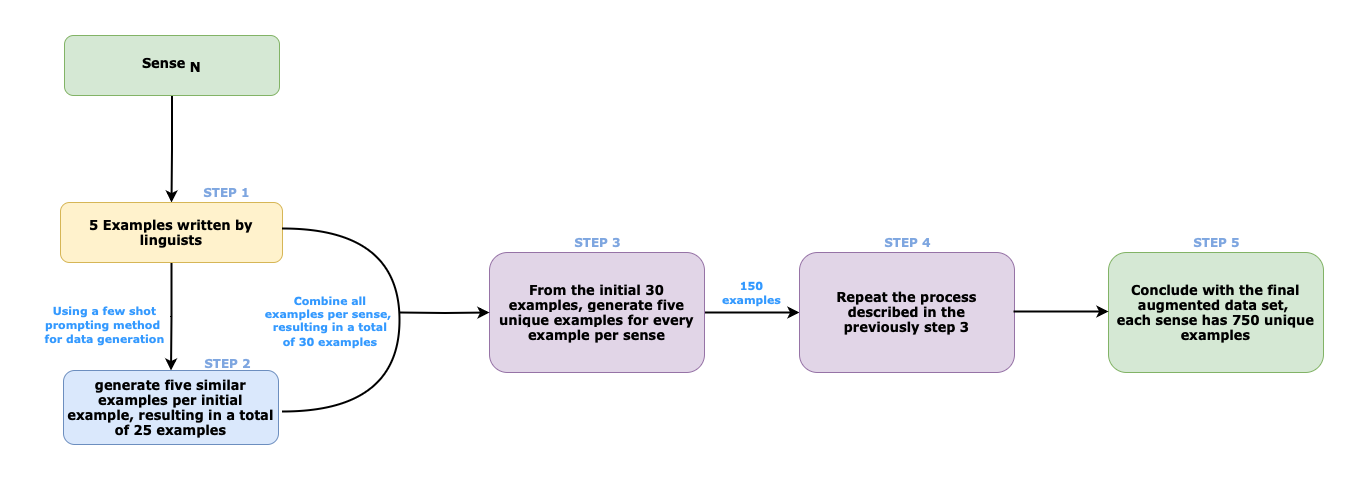}
}
\caption{Data generation and augmentation workflow for training examples.}
\label{Fig:3}
\end{figure*}

Synthetic data was necessary to produce training data for all these senses. This was because many senses are rare, infrequent, particularly the slang senses, which are not commonly used. In our test set, some senses appeared only a few times, which is evidence of how hard it is to find instances of these senses. 

 We created data using GPT-3.5-turbo \footnote{{\url{https://platform.openai.com/docs/models/}}} a large language model (LLM) developed by OpenAI with a prompt template designed to take a word, its available senses, and distinct examples for each sense to produce 750 examples per sense. The following template was used:

\begin{center}
     Template $=$ Word : \{word\} \\
    Sense: \{senses\} \\
    Example: \{unique\_examples\} \\
    \{instruction\}
\end{center}

The following steps describe the details of the data augmentation process:
\begin{itemize}
\item Step 1: Populate the template with five actual examples for each sense that linguists wrote.
\item Step 2: involves using a few-shot prompting method to generate five similar examples per initial example, resulting in a total of 25 examples.
\item Step 3: From the initial 30 sentences per sense, we re-inserted examples, words, and senses in the same template described above to create five new examples for each existing instance per sense, totalling 150 examples.
\item Step 4: Repeat the process described in Step 2 for each of the previously created twenty-five examples per sense. 
\item Step 5: concludes with the final augmented dataset, where each sense has 750 unique examples.\\
Figure \ref{Fig:3} explains the process for generating data based on relevant context.
\end{itemize}

\subsubsection{Human Evaluation Results}
\label{subsec:bdlsc-human-eval}

To evaluate the linguistic validity of the synthetic data used to train our supervised models, we conducted a structured human annotation study covering ten target words in the ST-WSD dataset. The same annotators who participated in the construction and validation of the BD-LSC and ST-WSD datasets conducted the evaluation, following the annotation guidelines described in Section~\ref{subsec:bdlsc_annotation}. We selected 200 examples per sense to balance representativeness and annotation feasibility, providing sufficient coverage of contextual variation while keeping the evaluation workload manageable. Because the number of senses per word ranges from two to eight, annotators rated 200 synthetic examples for every sense. Each example was evaluated on a five-point scale across three dimensions: \textit{Sense Accuracy}, \textit{Contextual Naturalness}, and \textit{Register/Genre Fidelity}. Sense Accuracy captures how clearly the example expresses the intended sense of the target word and whether the surrounding context supports an unambiguous interpretation. Contextual Naturalness reflects the grammaticality, idiomaticity, and semantic plausibility of the sentence. Register/Genre Fidelity measures how well the example matches the stylistic properties of the intended corpus, such as the historical tone of CCOHA-conditioned examples or the informal, conversational style typical of Twitter-conditioned examples.
\begin{itemize}[label=--]
\item \textit{\textbf{Sense-level evaluation}}
\label{app:bdlsc-sense-level-eval}

Each of the 200 examples per sense was rated independently by two annotators. For each evaluative dimension and each sense, we first computed an annotator-specific mean, defined as the average of that annotator’s 200 ratings. These two means were then averaged to obtain a combined sense-level score for Sense Accuracy, Contextual Naturalness, and Register/Genre Fidelity. Inter-annotator agreement was computed at the sense level using Cohen’s $\kappa$. To make the ratings suitable for $\kappa$, the five-point scale was collapsed to a binary decision: ratings of 4--5 were treated as acceptable, indicating that the example conveyed the intended meaning, sounded natural, and matched the expected register; ratings of 1--3 were treated as unacceptable, reflecting ambiguity, unnatural phrasing, or stylistic mismatch. These binary labels yielded two sets of 200 decisions per sense, from which Cohen’s $\kappa$ was calculated based on agreement about whether each example was acceptable or unacceptable.

\item \textit{\textbf {Word-level aggregation}}
\label{app:bdlsc-word-level-aggregation}

After computing sense-level scores, we derived word-level scores by averaging all sense-level results associated with each word. For example, if a word has six senses, the six sense-level means were averaged to yield a single Sense Accuracy score for that word; the same procedure was applied to Contextual Naturalness, Register/Genre Fidelity, and Cohen’s $\kappa$. Finally, the overall dataset means reported in Table~\ref{tab:human-eval} represent the average of the ten word-level values for each evaluation dimension, providing a global measure of the quality and reliability of the synthetic data.

\end{itemize}

\begin{table}[t]
\centering
\caption{Human evaluation results for synthetic examples across the ten target words.}
\begin{tabular}{lccccc}
\hline
\textbf{Word} & \textbf{\# Senses} & \textbf{Sense Acc.} & \textbf{Naturalness} & \textbf{Register} & \textbf{Cohen's $\kappa$} \\
\hline
Eat        & 6 & 4.6 & 4.5 & 3.7 & 0.72 \\
BMW        & 3 & 4.4 & 4.2 & 3.5 & 0.69 \\
Brownie    & 8 & 4.7 & 4.6 & 3.8 & 0.72 \\
Chronic    & 5 & 4.5 & 4.3 & 3.6 & 0.68 \\
Climber    & 4 & 4.4 & 4.1 & 3.4 & 0.67 \\
Germ       & 6 & 4.6 & 4.4 & 3.5 & 0.71 \\
Mammy      & 4 & 4.8 & 4.6 & 3.9 & 0.72 \\
Cucumber   & 3 & 4.5 & 4.3 & 3.4 & 0.68 \\
Rodent     & 2 & 4.3 & 4.0 & 3.3 & 0.66 \\
Salty      & 6 & 4.7 & 4.5 & 3.7 & 0.72 \\
\hline
\textbf{Overall Mean} & -- & \textbf{4.55} & \textbf{4.35} & \textbf{3.58} & \textbf{0.70} \\
\hline
\end{tabular}
\label{tab:human-eval}
\end{table}



\subsection{Unsupervised Approach}
\label{subsec:bdlsc_unsupervised}


\subsubsection{Clustering on Manifolds of Contextualised Embeddings for WSD}
\label{subsubsec:clustering_manifolds_wsd}

 Our study uses ALBERT-xxlarge-v2 contextualised word embeddings~\citep{lan2019albert} for word meaning representation. Contextualised embeddings enable tracking finer sense variations in words across time~\citep{hu2019diachronic}. We have utilised the ~\say{sentence-transformers}; these embeddings are particularly effective for document-level tasks, providing accurate and meaningful representations of text data. We calculate contextualised embedding vectors with a dimensionality of \texttt{4096} for each focus word within its sentence context. The underlying idea is that senses or meanings are represented by closely related vectors in terms of cosine distance since they appear in similar contexts. We utilise the ALBERT-xxlarge-v2 model from Huggingface~\footnote{\url{https://github.com/google-research/ALBERT}}, which consists of \texttt{223M} parameters~\citep{lan2019albert}. The attractiveness of Albert lies in several key factors ~\citep{lan2019albert}: 
\begin{enumerate}
 \item Reported strong performance, competitive with or exceeding BERT on several NLP benchmarks~\citep{lan2019albert}.
    \item Albert incorporates two parameter reduction techniques (factorised embedding parametrisation and cross-layer parameter sharing) that lift the major obstacles in scaling pre-trained models and improve parameter efficiency. 
    \item Albert incorporates a self-supervised loss for sentence-order prediction (SOP). This aspect primarily emphasises inter-sentence coherence and aims to rectify the shortcomings identified in the original BERT's proposed next-sentence prediction (NSP) loss.
\end{enumerate}
\begin{figure*}[!ht]
    \centering
\captionsetup{justification=raggedright,singlelinecheck=false} 
    \includegraphics[width=0.9\textwidth]{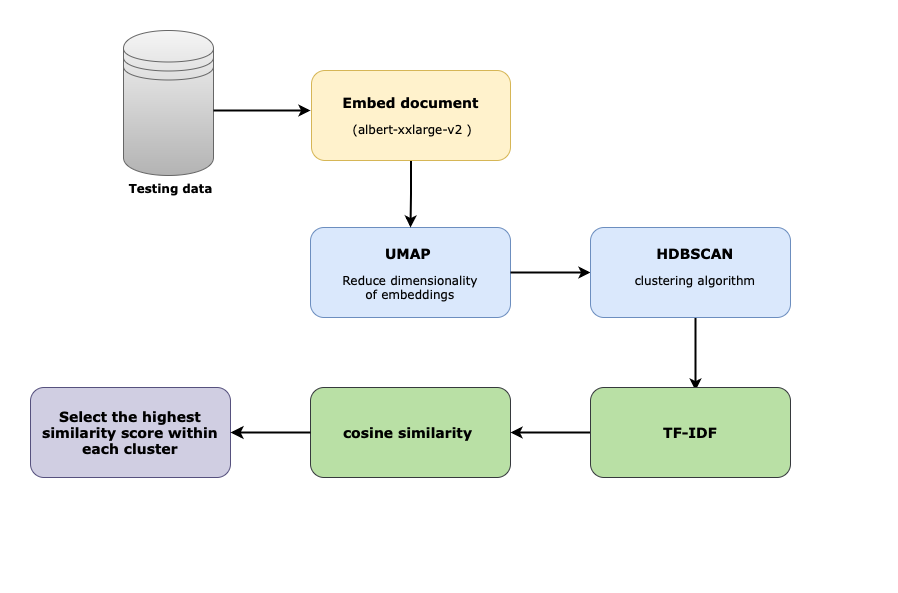} 
    \caption{Unsupervised WSD pipeline based on contextualised embeddings and clustering.}
    \label{Fig:5}
\end{figure*}


Subsequently, we employ dimensionality reduction techniques. It's essential to reduce the dimensionality of the embeddings because numerous clustering algorithms struggle when dealing with high-dimensional data, and dimensionality reduction enhances the overall stability of the system~\citep{mcinnes2018umap}. We employ the UMAP algorithm, a form of manifold learning ~\citep{mcinnes2018umap}. 
For our final UMAP hyperparameter configuration, we have chosen the same configuration as in the CMCE paper~\citep{rother2020cmce}, which uses six neighbours. The number of neighbours plays a crucial role as it determines how many points the algorithm associates with each point in the embedding manifold. We have also set the minimal distance between points in the embedding space to \texttt{0.0}. This minimal distance should typically be greater than zero for visualisation purposes only. Upon diminishing the dimensionality of the document embeddings to \texttt{6}, we are able to cluster the documents using HDBSCAN~\citep{mcinnes2017hdbscan}. HDBSCAN is an algorithm that works effectively with UMAP due to its ability to preserve a lot of local structures even when reduced to a lower-dimensional space.

 \begin{figure}
    \centering
    \includegraphics[width=0.8\linewidth]{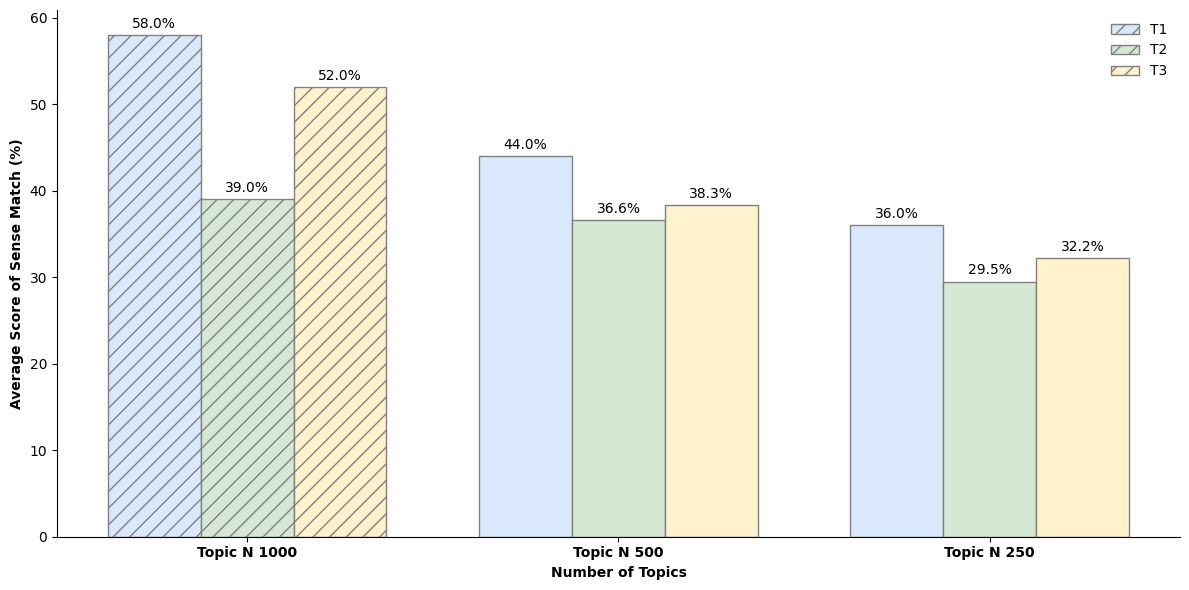}
    \caption{Comparison of results with different numbers of top words. The highest values in each time period are highlighted with a hatched pattern.}
    \label{fig:4}
\end{figure}


\subsubsection{Semantic Change Detection}
\label{subsubsec:semantic_change_detection}

To derive topics from each cluster, we have utilised Term Frequency-Inverse Document Frequency (TF-IDF) to identify frequently occurring topics within each group of documents compared to others. The result of the TF-IDF score would highlight the significant words within a given topic. To create a topic representation, we select the top \texttt{1000} words for each topic, determined by their TF-IDF scores, with higher scores indicating a greater significance of the word to its topic due to the score reflecting information density. We experimented with different top word settings to conclude by choosing the top 1000 words. Figure~\ref{fig:4} shows a comparison of the average sense match percentages of all time periods between the top 250, 500 words, and 1000 words. We can see that the results with the top 1000 words achieve the best sense match percentages in all timelines. After finalising our senses, we compared them across each timeline and calculated the LSCD accuracies by analysing the differences between the three time periods.

 We then compute the cosine similarity between the definitions' embedding and the embeddings of the top \texttt{1000} words in each topic (cluster). The highest similarity score within each cluster was selected. The schematic representation of this model is presented in Figure~\ref{Fig:5}.



\subsection{Supervised Machine Learning}
\label{subsec:bdlsc_supervised_ml}

We describe our systems for Lexical Semantic Change Detection in English. To model change detection, we approach the task by formulating it as a WSD problem. 

\begin{figure}[!ht]
    \centering
    \includegraphics[width=0.8\columnwidth]{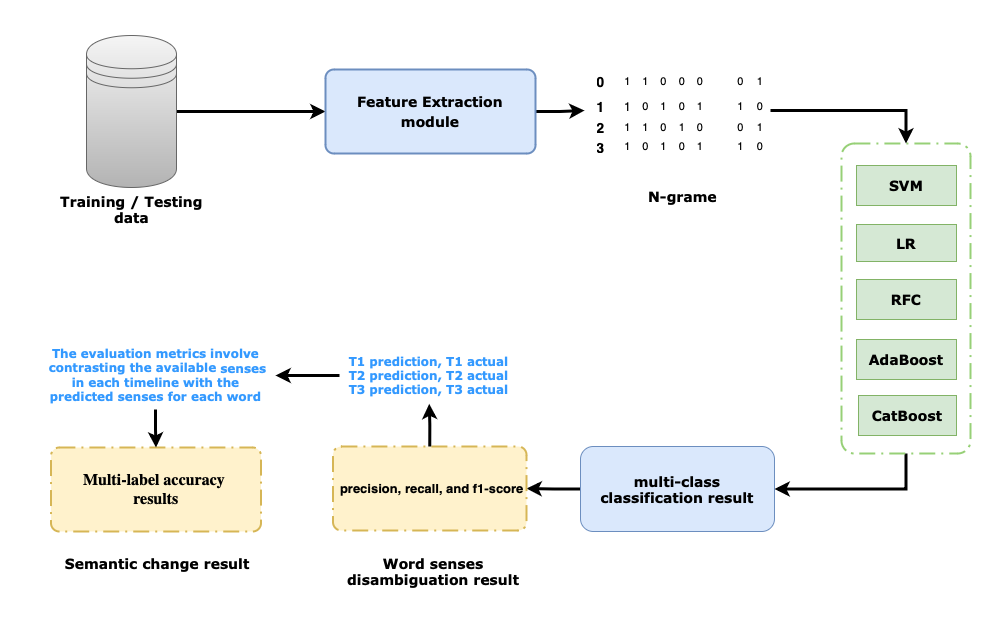}
    \caption{Supervised machine-learning pipeline for WSD and LSCD.}
    \label{fig:6}
\end{figure}

We build a set of baseline classifiers, including traditional ML algorithms such as Random Forest (RF)~\citep{breiman2001random}, Logistic Regression (LR)~\citep{menard2002applied}, Support Vector Machines (SVM)~\citep{hearst1998support}, Adaptive Boosting (AdaBoost)~\citep{Freund_AdaBoostM1}, and Category Boosting (CatBoost). These models are used for the WSD task. The parameters for these models are presented in Table~\ref{tab:model_hyperparameters}.

We used the following features in our experiments:  
\begin{enumerate}
    \item N-grams~\citep{sidorov2014syntactic}: Word N-grams ($n = 1,2,3$), character N-grams ($n = 1,2,3$), and various combinations, including word (1,2) grams, character (1,2) grams, word (1,3) grams, and character (1,3) grams, applied across all supervised algorithms.  
    \item DistilBERT embeddings (distilbert-base-uncased)~\citep{sanh2019distilbert}: Used with the best-performing algorithm (RF) in experiments conducted with N-grams.  
    \item FastText embeddings (wiki-news-300d-1M.vec): A set of one million word vectors trained on Wikipedia 2017, the UMBC webbase corpus, and the statmt.org news dataset (16B tokens)~\citep{grave2018learning}, also used with the best-performing algorithm (RF) in experiments conducted with N-grams.  
\end{enumerate}

Once the classifiers had labelled word senses for each time period, we compared results across time periods to determine LSCD accuracies.  

The overview of the proposed model is shown in Figure~\ref{fig:6}, while the feature extraction module is presented in Figure~\ref{Fig:7}. The hyperparameters utilised for these models are provided in Table~\ref{tab:model_hyperparameters}.

\begin{figure}[!hbpt]
    \resizebox{\columnwidth}{!}
  {%
\includegraphics[width=\textwidth]{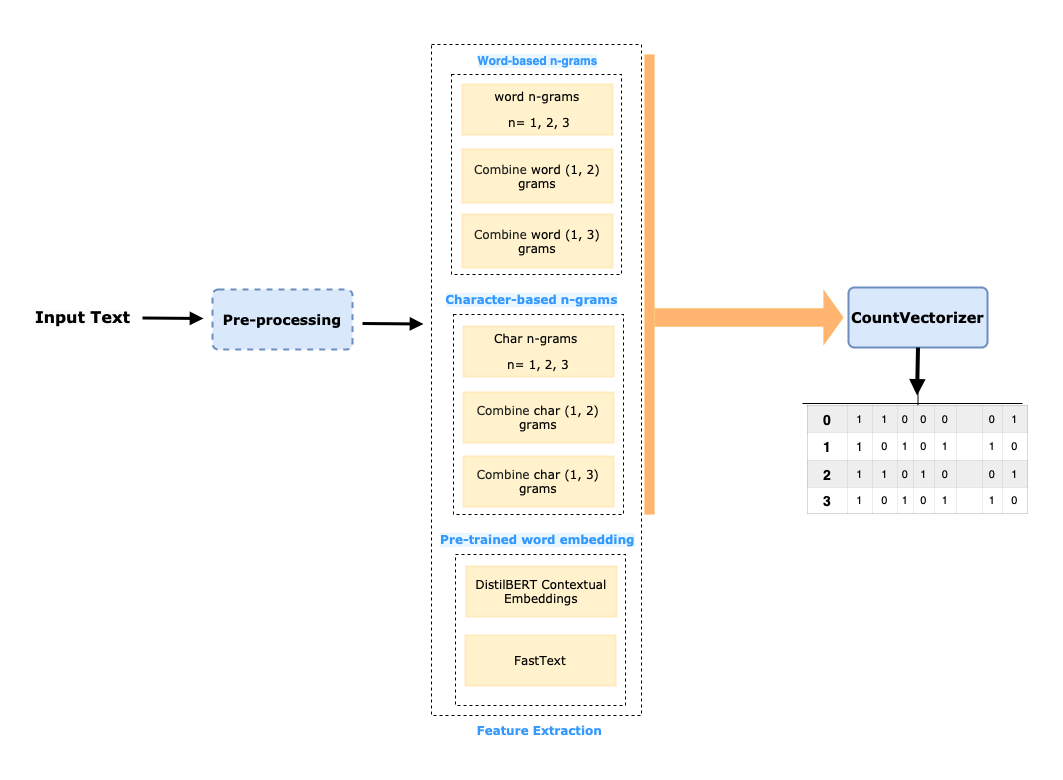}

}

\caption{Feature extraction module used in supervised WSD models.}
\label{Fig:7}

\end{figure}


\subsection{Transformer-based Models}
\label{subsec:bdlsc_transformers}

We employed two large transformer-based models, BERT-large~\citep{devlin2019bert} and RoBERTa-large~\citep{liu2019roberta}, to perform word sense disambiguation across the three time periods. Both models were fine-tuned using five-fold cross-validation for 30 epochs. Although trained under identical optimisation settings, the two architectures differ in important ways. BERT-large uses a bidirectional masked language modelling objective, whereas RoBERTa-large extends BERT by removing next-sentence prediction, training on significantly larger corpora, and using optimised hyperparameters. These distinctions enable RoBERTa-large to capture contextual nuances more effectively in many downstream tasks.

The hyperparameters used for both models are listed in Table~\ref{tab:model_hyperparameters}. After fine-tuning, each model produced sense labels for every instance in each time period. We then compared the predicted senses across pairs of periods to compute LSCD accuracies.



\subsection{Large Language Models (LLMs)}
\label{subsec:bdlsc_llms}

In this study, we leverage the OpenAI models GPT-4o and GPT-4o-mini to evaluate their performance using two distinct prompting strategies: zero-shot and few-shot prompting~\citep{brown2020language}. Zero-shot prompting enables the models to generate responses without prior exposure to specific examples, thereby testing their ability to generalise solely from the task description. In contrast, few-shot prompting incorporates carefully curated examples within the input prompt, enhancing the model’s contextual understanding and performance by providing guidance on the desired format and response style.  

After the models assigned word senses for each time period, we analysed and compared the results across different time periods to evaluate the accuracy of LSCD. The parameters and prompts used with these models are presented in Table~\ref{tab:model_hyperparameters} and Tables~\ref{tab:zero-shot-prompting} and~\ref{tab:few-shot-prompting}.


\section{Results}
\label{sec:bdlsc_results}

Table~\ref{tab:wsd-performance} provides a detailed comparison of the performance of several approaches to word sense disambiguation across the three time periods. The N-gram and Character N-gram model shows consistent accuracy, with values of 0.66 in T1, 0.68 in T2, and 0.68 in T3. Although the accuracy remains stable, the low Macro F1 values indicate a limited ability to recognise less frequent senses and to manage variation across classes.

The DistilBERT model records noticeably lower performance in all periods. Both accuracy and F1 scores remain at modest levels, reflecting an overall difficulty in capturing the contextual information required for reliable sense interpretation. FastText shows slightly higher results than DistilBERT, particularly in Micro F1, yet the Macro F1 values remain limited. This pattern reflects the constraints of relying on static representations that do not fully capture the evolving meanings of words across contexts.

The transformer-based models provide a substantial improvement over the earlier approaches. BERT large achieves strong accuracy in each period, reaching 0.72 in T1, 0.75 in T2, and 0.77 in T3. RoBERTa large performs at a comparable level, though with slightly lower accuracy. These models benefit from their ability to incorporate fine-grained contextual information, which improves their capacity to distinguish between subtle sense variations.

The GPT-4o models deliver the strongest overall results. Both the zero-shot and few-shot configurations perform well across all periods, with the few-shot configuration achieving the highest accuracy. The few-shot GPT-4o model reaches 0.77 in T1, 0.80 in T2, and 0.83 in T3. The continued improvement across periods indicates that the inclusion of a small set of examples enables the model to refine its interpretation of ambiguous contexts and to identify word senses with greater precision.


\begin{table*}[t]
\centering
\caption{Comparison of WSD performance across various methods.
The results represent average performance metrics evaluated on ten target words across three distinct time periods (T1, T2, and T3). 
Bold Accuracy values indicate the best results, while underlined Accuracy values represent the second-best results.}
\renewcommand{\arraystretch}{1.2}
\setlength{\tabcolsep}{10pt}
\resizebox{\textwidth}{!}{%
\begin{tabular}{c ccc ccc ccc}
\toprule
\textbf{Method} &
\multicolumn{3}{c}{\textbf{T1}} &
\multicolumn{3}{c}{\textbf{T2}} &
\multicolumn{3}{c}{\textbf{T3}} \\
\cmidrule(lr){2-4} \cmidrule(lr){5-7} \cmidrule(lr){8-10}
& \textbf{Accuracy} & \textbf{Macro-F1} & \textbf{Micro-F1}
& \textbf{Accuracy} & \textbf{Macro-F1} & \textbf{Micro-F1}
& \textbf{Accuracy} & \textbf{Macro-F1} & \textbf{Micro-F1} \\
\midrule

\textbf{N-gram and Character N-gram ML}  & 0.66 & 0.32 & 0.66 & 0.68 & 0.31 & 0.67 & 0.68 & 0.32 & 0.66 \\
\textbf{DistilBERT-based ML}             & 0.33 & 0.20 & 0.33 & 0.39 & 0.25 & 0.44 & 0.47 & 0.30 & 0.53 \\
\textbf{FastText-based ML}               & 0.36 & 0.25 & 0.40 & 0.39 & 0.27 & 0.42 & 0.47 & 0.31 & 0.53 \\
\midrule

\textbf{BERT-large}                      & 0.72 & 0.50 & 0.72 & 0.75 & 0.55 & 0.75 & 0.77 & 0.66 & 0.77\\
\textbf{RoBERTa-large}                   & 0.69 & 0.47 & 0.69 & 0.72 & 0.52 & 0.72 & 0.73 & 0.63 & 0.74 \\
\midrule

\textbf{GPT-4o-mini zero-shot}           & \underline{0.76} & 0.48 & 0.76 & 0.77 & 0.58 & 0.77 & 0.81 & 0.56 & 0.81 \\
\textbf{GPT-4o zero-shot}                & \underline{0.76} & 0.55 & 0.76 & \underline{0.78} & 0.45 & 0.78 & \underline{0.82} & 0.44 & 0.82 \\
\textbf{GPT-4o-mini few-shot}            & 0.74 & 0.53 & 0.74 & 0.75 & 0.54 & 0.75 & 0.80 & 0.58 & 0.80 \\
\textbf{GPT-4o few-shot}                 & \textbf{0.77} & 0.49 & 0.77 & \textbf{0.80} & 0.54 & 0.80 & \textbf{0.83} & 0.54 & 0.83 \\

\bottomrule
\end{tabular}
}
\label{tab:wsd-performance}
\end{table*}

Table~\ref{tab:Comparison of Exact Sense Match and Multi-Label Accuracy} presents the results of Exact Sense Match (ESM) and Multi-Label Accuracy (MA) across the evaluated methods. The unsupervised approach, ALBERT-xxlarge-v2, demonstrates lower accuracy across all time periods, achieving ESM scores of 58.31\% in T1, 39.99\% in T2, and 52.66\% in T3. Its multi-label accuracy is likewise low and uneven across period pairs (53.1\% for T1--T2, 66.0\% for T2--T3, and 43.1\% for T1--T3), indicating that this approach struggles to capture complex lexical variation and evolving slang meanings.

Our findings suggest that the unsupervised clustering approach evaluated here is insufficient for detecting semantic shifts in slang usage. We do not claim this generalises to all unsupervised methods; rather, it reflects a known limitation of distribution-based clustering, which, as noted by~\citep{schlechtweg2020semeval}, identifies change by comparing dominant sense distributions across time and is therefore sensitive to frequent, stable meanings rather than newly emerging ones. In our setting this limitation manifests through several underlying factors:
\begin{enumerate}
    \item The model predominantly identifies dominant word senses, performing well when standard meanings are well represented in the dataset. However, it struggles to predict slang senses, which often deviate from conventional linguistic patterns. Because unsupervised models lack explicit training signals, they tend to reinforce the most frequent word senses rather than capture emerging, context-dependent meanings.

    \item Slang senses are often infrequent in the dataset, making them difficult to predict without direct supervision. Unlike standard word senses that appear consistently, slang meanings evolve rapidly and exhibit high variability, leading to data sparsity issues. Because unsupervised models lack explicit guidance, they struggle to generalise well in such cases. This limitation motivated our transition to supervised methodologies, which provide a more structured learning framework for capturing diverse semantic representations.

    \item While unsupervised methods perform well in binary classification tasks where a word has two distinct, well-defined senses, their effectiveness diminishes when words exhibit multiple or overlapping meanings. In binary cases, distinguishing between two standard senses is relatively straightforward, but slang words often display fluid, intersecting senses that pose challenges for unsupervised models.
\end{enumerate}

\begin{table*}
\centering
\caption{Comparison of Exact Sense Match and multi-label accuracy performance across various methods. 
The results represent average performance metrics evaluated on ten target words across three distinct time periods (T1, T2, and T3). Bold values indicate the best results, while underlined values represent the second-best results for each metric.}
\renewcommand{\arraystretch}{1.2}
\setlength{\tabcolsep}{8pt}
\resizebox{\textwidth}{!}{%
\begin{tabular}{l ccc ccc}
\toprule
\textbf{Method} & \multicolumn{3}{c}{\textbf{Exact Sense Match (\%)}} & \multicolumn{3}{c}{\textbf{Multi-Label Accuracy (\%)}} \\ 
\cmidrule(lr){2-4} \cmidrule(lr){5-7}
& \textbf{T1 Avg} & \textbf{T2 Avg} & \textbf{T3 Avg} & \textbf{T1--T2 Avg} & \textbf{T2--T3 Avg} & \textbf{T1--T3 Avg} \\ 
\midrule

\textbf{ALBERT-xxlarge-v2 (unsupervised)} 
& 58.31 & 39.99 & 52.66 & 53.1 & 66.6 & 43.1 \\

\textbf{N-gram and Character N-gram ML}
& 79.65 & \underline{86.66} & 80.81 & 66.5 & \textbf{79.9} & 66.6 \\

\textbf{DistilBERT-based ML} 
& 76.81 & 71.70 & 78.49 & 63.0 & 76.5 & \underline{83.3}\\

\textbf{FastText-based ML} 
& 71.48 & 71.65 & 65.16 & 46.6 & 63.1 & 73.1 \\
\midrule

\textbf{BERT-large} 
& \underline{85.99} & 77.14 & \underline{89.57} & 63.1& \underline{76.6} &79.9 \\

\textbf{RoBERTa-large} 
& 72.65 & 74.97 & 88.32 & 46.4 & 73.2 & 69.9 \\
\midrule

\textbf{GPT-4o-mini zero-shot} 
& 75.98 & 80.47 & 84.98 & \underline{69.9}& \textbf{79.9} & 83.2\\

\textbf{GPT-4o zero-shot} 
& \underline{85.99} & 79.14 & \underline{89.57} & 63.1 & \underline{76.6} &\textbf{ 86.6} \\

\textbf{GPT-4o-mini few-shot} 
& 77.65 & 80.47 & 89.15 & 63.1 & \underline{76.6} & 79.9 \\

\textbf{GPT-4o few-shot} 
& \textbf{87.99} &\textbf{89.65}  & \textbf{90.82} & \textbf{76.4} & \underline{76.6} & \textbf{86.6} \\

\bottomrule
\end{tabular}
}
\label{tab:Comparison of Exact Sense Match and Multi-Label Accuracy}
\end{table*}

Examining the results of DistilBERT and FastText-based machine learning models reveals a clear performance gap between these embedding-based approaches and traditional N-gram methods. The N-gram models consistently achieve higher averages for both ESM and MA, demonstrating their robustness in capturing explicit lexical variations. In contrast, DistilBERT and FastText exhibit weaker performance, particularly in cases of lexical ambiguity where multiple senses coexist within similar contextual environments.

The comparatively weaker performance of embedding-based transformer models, such as DistilBERT, can be attributed to their highly contextual nature. While this property is advantageous for many NLP tasks, it introduces specific challenges for word sense disambiguation. When used as static or feature-based embeddings, transformer representations are conditioned on surrounding context, which can blur distinctions between ambiguous senses and make it more challenging to separate multiple meanings. 

Contextualised transformer models outperform static or shallow embedding approaches because they can model fine-grained contextual semantics. Prior work has shown that contextualised embeddings capture meaning variation more effectively than static representations ~\citep{martinc2020leveraging, laicher2021explaining}, and our results reflect this pattern: fine-tuned BERT-large and RoBERTa-large achieve substantially higher and more consistent performance than FastText, DistilBERT-based machine learning, or N-gram baselines.

Our findings align with recent studies that emphasise the 
importance of few-shot prompting for large language model based 
lexical semantic change detection. Consistent 
with~\citep{periti2024chat}, GPT-4o shows substantial improvement 
when given few-shot examples. In our experiments, GPT-4o few-shot 
prompting provides the strongest performance across Exact Sense 
Match and Multi-Label Accuracy. For Exact Sense Match, GPT-4o 
few-shot achieves the highest scores in all three time periods, 
reaching 87.99 percent in T1, 89.65 percent in T2, and 90.82 
percent in T3. These values exceed the results of all 
transformer-based models and all machine learning baselines. A 
similar pattern appears in the Multi-Label Accuracy results, where 
GPT-4o few-shot attains the highest accuracy for T1 to T2 with 
76.4 percent, for T2 to T3 with 76.6 percent, and for T1 to T3 
with 86.6 percent. The N-gram and Character N-gram model produces a slightly higher accuracy for T2 to T3 (79.9 percent), so GPT-4o few-shot is not strictly best in every interval. On aggregate, however, it is the most consistent performer across Exact Sense Match and multi-label accuracy, a reading we qualify below in light of the Macro-F1 scores. The complete results are presented in Tables~\ref{tab:wsd_results_full}.

A more cautious reading emerges from the Macro-F1 scores. Whereas accuracy, Micro-F1, and ESM are dominated by the frequent standard senses, Macro-F1 weights every sense equally and therefore reflects performance on the rare slang senses that motivate this study. On this metric no system exceeds 0.66, and the best-performing system overall, GPT-4o few-shot, attains only 0.49--0.54 across the three periods, below fine-tuned BERT-large at T3 (0.66). This gap indicates that the high accuracy and ESM values are driven largely by correctly resolving common standard senses, while all systems, including the strongest, continue to misclassify a substantial proportion of low-frequency slang senses. We therefore present GPT-4o few-shot as the most consistent system on aggregate sense coverage rather than as a solution to slang-sense disambiguation, which remains an open problem. Per-word ESM and MA breakdowns are given in Appendix~\ref{chap:appendix-a-bdlsc} (Tables~\ref{tab:Exact sense match result} and~\ref{tab:multi-labelled-accuracy}).


\section{Error Analysis and Discussion}
\label{sec:bdlsc_discussion}

To systematically evaluate the limitations of the best-performing models across different methodological categories, we conducted a detailed error analysis. The selected models represent the top performers in each category: the N-gram and Character N-gram model from traditional machine learning, the fine-tuned BERT-large transformer model, and the GPT-4o few-shot model from large language models (LLMs).

A total of 100 misclassified examples were sampled from each model's prediction set and manually analysed to identify recurring patterns and sources of error.\footnote{For readability, the examples in this section are shown in their original form, prior to the lowercasing and username removal described in Section~\ref{subsec:bdlsc_preprocessing}.} These errors were grouped into distinct linguistic and contextual categories to enable a structured comparison of model behaviour and failure modes.


\subsection{Context Understanding (N-gram: 24\%, BERT-large: 20\%, GPT-4o few-shot: 18\%)}
\label{subsec:context_understanding_results}

All three models demonstrated difficulties in interpreting contextual information, with errors occurring more frequently in the N-gram model. For example, in the sentence \say{Brownie crept along the hallway, then bolted the moment the front door opened}, the N-gram model classified \textit{Brownie} as food, ignoring contextual cues such as \say{crept} and \say{bolted}, which indicate that \textit{Brownie} is being used as a proper name (S7). In contrast, BERT-large correctly disambiguated the intended meaning, as did GPT-4o.

However, a similar issue arose with the word \textit{rodent}. In the sentence \say{The villagers set traps overnight, but the rodent kept returning to steal their food}, the correct sense is the literal animal, supported by contextual indicators such as \say{traps} and \say{food}. The N-gram model correctly identified the literal sense. In contrast, BERT-large and GPT-4o both misclassified it as the slang insult sense when they failed to attend closely to the immediate context. Notably, here the simpler N-gram model succeeded where both neural models failed, showing that the general contextual advantage of transformer and large language models is not uniform and can reverse when the disambiguating cue is local and lexically explicit.

These results indicate that although transformer models and large language models are generally more contextually robust than traditional approaches, all can fail when contextual signals are weak or under-represented in the training data. Such misclassifications demonstrate that even strong models may overlook relevant semantic cues when they appear outside dominant or expected patterns.


\subsection{Handling Named Entities (N-gram: 12\%, BERT-large: 10\%, GPT-4o few-shot: 9\%)}
\label{subsec:handling_named_entities_results}

Named entities were another primary source of errors across models. The N-gram model had particular difficulty distinguishing proper nouns from common nouns. For example, in the sentence \say{Brownie was able to walk by the time they left the field}, the N-gram model classified \textit{Brownie} as a generic noun, ignoring its use as a proper name. 

The BERT-large model showed similar difficulties but for different reasons. Because BERT relies heavily on distributional priors from large-scale training data, it often defaulted to the most frequent or canonical meaning of an entity. For instance, in the sentence \say{Chronic has been on repeat ever since I got tickets yesterday}, BERT-large incorrectly selected the medical sense of \textit{Chronic} rather than recognising it as the title of Dr. Dre’s album. This reflects the model’s tendency to favour high-frequency senses and to overlook culturally specific or low-frequency named entities.

GPT-4o also showed occasional errors with named entities, particularly when interpretation depended on cultural or contextual cues. In the tweet \say{Ever since I moved to Seattle, I’ve had ‘Chronic’ on repeat, the beats keep bringing me back to those unforgettable concerts back home! \#DrDre \#Throwback}, GPT-4o interpreted \textit{Chronic} as cannabis instead of recognising it as the title of Dr.~Dre’s album. The N-gram model made a similar mistake, defaulting to the medical sense and failing to identify the named entity.


\subsection{Long Range Dependencies (N-gram: 15\%, BERT-large: 14\%, GPT-4o few-shot: 13\%)}
\label{subsec:long_range_dependencies_results}

Interpreting sentences with dependencies spread across extended contexts posed challenges for all models, particularly when they relied heavily on localised patterns or surface-level cues. The N-gram model struggled with references that required connecting distant contextual information. For example, in the sentence \say{Brownie kept him company when she went to work, showing him how to do simple household chores and telling him stories}, the N-gram model classified \textit{Brownie} as a dessert, failing to use later contextual clues that indicate \textit{Brownie} is a character. 

The BERT-large model showed improved performance by capturing some long-range relationships within sentences, yet it occasionally failed when essential disambiguating information appeared across multiple clauses. In such cases, the model sometimes focused on more local lexical cues rather than integrating the entire context.

A similar issue appeared with the word \textit{salty}. In the tweet \say{He snapped at me in the group chat, salty at first lol but later I realised he was just joking, no harm done \#chill}, the correct sense is the slang meaning “irritated or annoyed.” Initially, the phrase may suggest the literal taste sense of \textit{salty}, but the later cue \say{he was just joking} clarifies the slang interpretation. Both the N-gram and BERT-large models correctly identified the slang sense by attending to these later contextual cues. In contrast, GPT-4o misclassified the word as the literal taste sense when it failed to integrate information from the full sentence context. This is a further case in which the expected capability ordering inverts, with the two weaker models succeeding where the strongest model did not.


\subsection{Idioms and Common Phrases (N-gram: 7\%, BERT-large: 10\%, GPT-4o few-shot: 14\%)}
\label{subsec:idioms_common_phrases_results}

Understanding idioms and common phrases proved challenging for all models, although GPT-4o faced greater difficulty overall. For instance, GPT-4o struggled with \say{Stop inviting that germ to every party, he ruins the mood the second he walks in}, interpreting \textit{germ} literally as a microorganism rather than recognising the insult sense (S4), in which a person is described as contemptible or unpleasant. In contrast, the N-gram model handled this case more accurately by identifying the figurative usage.

However, the N-gram model was not consistently better at idiomatic interpretation and often failed with clearly figurative expressions. Consider the sentence \say{Jake’s been making serious moves in the underground rap scene, he eats up every opportunity that comes his way. \#GrindNeverStops}. The N-gram model completely missed the mark, interpreting \textit{eat} literally as the act of consuming food rather than recognising its slang meaning of eagerly taking advantage of opportunities.



\subsection{Metaphors and Sarcasm (N-gram: 12\%, BERT-large: 10\%, GPT-4o few-shot: 9\%)}
\label{subsec:metaphors_sarcasm_results}

Metaphors and sarcasm were particularly challenging for all models, although GPT-4o generally handled them better overall. The N-gram model tended to interpret metaphors too literally, often missing their intended figurative meanings. For example, in the tweet \say{Watch out for her, she's a real climber in London's high society. She’ll step on anyone to reach the top. \#SocialClimber \#LondonLife}, the N-gram model incorrectly assumed that the word \textit{climber} referred to someone physically climbing. In reality, the phrase metaphorically describes a person determined to advance socially, often by disregarding others.

The BERT-large model, while more contextually aware, struggled with recognising sarcasm when explicit tonal cues were limited. The phrase \say{I didn’t know that rodent intelligence has progressed as far as mad gaming skills} is clearly sarcastic, mocking the subject through exaggerated praise. BERT-large failed to identify the sarcastic intent and instead interpreted the statement literally, associating it with an actual animal reference rather than irony. 

Despite its overall stronger understanding of context, GPT-4o also misinterpreted sarcasm in subtle cases. For instance, when faced with the statement \say{Great idea, let's ignore the warnings and eat sushi from the gas station, what could possibly go wrong?}, it treated the expression as genuine agreement rather than recognising the clear ironic tone. Both models, despite their advanced linguistic capabilities, continue to struggle with nuanced figurative and sarcastic expressions that depend on pragmatic understanding and real-world inference.


\subsection{Unknown Errors (N-gram: 30\%, BERT-large: 36\%, GPT-4o few-shot: 37\%)}
\label{subsec:unknown_errors_results}

A considerable proportion of misclassifications across all models fell into the category of unknown errors, where mistakes could not be clearly linked to a specific linguistic challenge. Unlike issues related to context understanding, named entities, or figurative language, these cases occurred even when sufficient contextual information was present to guide correct interpretation. This pattern suggests that despite substantial progress in modern language models, fundamental limitations remain in how systems prioritise competing word senses and handle subtle lexical ambiguity.  

One example is \say{My BMW played lil' dude too ... BMW = Black Mexican White \#pimpin}, which contains a rare, community-specific slang reinterpretation of the acronym \textit{BMW}. All models, including BERT-large, defaulted to the dominant \say{car} sense, reflecting the challenge of handling niche or emergent usages rather than systematic issues such as polysemy or idiomatic misunderstanding.

A further illustration is \say{Ugh, I pulled an all-nighter, drank way too much cheap coffee, and now my mouth tastes like rodent death. Never again lol.}, an exaggerated figurative use expressing disgust. The N-gram and BERT-large models produced literal interpretations, while GPT-4o also failed to infer the intended hyperbole. Similarly, in the historical example \say{Their clothes came from a fashionable modiste, but I always said, “Oh, this is a little something Mammy ran up for me.” So when I walked into the great hall at Winston … Mammy wasn’t no mammy to it. Mammy was a servant, a slave who had nursed the white woman.}, all models defaulted to the generic \say{mother} sense, overlooking the socio-historical connotation of \textit{mammy} as a racialised stereotype rooted in slavery.  

These unknown errors reveal that while transformer and large language models capture context more effectively than traditional approaches, they still exhibit unpredictable failures when processing unconventional, culturally specific, or pragmatically irregular expressions. Overcoming such challenges will require advances in discourse-level reasoning, social context modelling, and adaptive meaning prioritisation to achieve a more human-like understanding of linguistic nuance.



\section{Conclusion}
\label{sec:bdlsc_conclusion}

This paper presented the first benchmark and systematic evaluation for detecting semantic change in words that exhibit both slang and standard meanings. To address the lack of resources capable of capturing multi-directional change, we introduced two new datasets: BD-LSC, which traces sense gain, sense loss, and stability across three time periods, and ST-WSD, which provides fine-grained instance-level annotations for words combining slang and standard senses. Using these datasets, we systematically evaluated unsupervised, supervised, transformer-based, and large language models. Our results show that unsupervised clustering struggles with slang-driven and infrequent senses, largely because it relies on dominant sense distributions. Supervised machine learning models provide more stable performance but remain limited by shallow contextual representations. Fine-tuned transformer models demonstrate stronger contextual understanding, particularly for overlapping and subtle senses. GPT-4o in the few-shot setting achieves the strongest aggregate performance, with the highest Exact Sense Match scores and the best multi-label accuracy in two of the three evaluation intervals. However, Macro-F1 scores close to 0.5 across all systems indicate that low-frequency slang senses remain poorly handled, so this result should be read as a strong baseline rather than a solved task. Together, these findings highlight the importance of contextual modelling and task-specific supervision for capturing complex semantic trajectories. The BD-LSC and ST-WSD datasets establish a foundation for future work on slang evolution and bi-directional semantic change, enabling more robust and linguistically grounded evaluation of semantic change detection systems.



\section{Limitations}
\label{sec:bdlsc_limitations}
This work focussed on a deliberately narrow set of ten target words 
for the instance-level ST-WSD dataset. As a result, we did not extend 
the fine-grained sense annotation to a larger inventory of words or 
parts of speech. The primary motivation for this consideration is the 
annotation burden of the task, as each occurrence required full manual 
sense assignment across three time periods, with paragraph-level CCOHA 
contexts that frequently place the disambiguating cue several sentences 
away from the target word; restricting the scope in this way follows 
established practice in semantic change research, where controlled 
pilot datasets precede broader scaling~\citep{schlechtweg2020semeval}. 
Also, as we mentioned previously, the three periods are drawn from 
corpora that differ in register, namely CCOHA for the historical 
periods and Twitter for the contemporary one, so some apparent change 
is register-conditioned rather than purely temporal; we therefore read 
change at the sense level rather than from raw frequency.



\newpage
\appendix

\section{Corpus Overview}
\label{chap:appendix-a-bdlsc}
\label{app:bdlsc-corpus-overview}

 The table~\ref{tab:Overview-of-Corpus} offers a detailed view of the corpus, focusing on a list of target words and their total number of senses. It also includes frequency data for these target words.


\renewcommand{\arraystretch}{1.1}

\begingroup
\small 
\begin{longtable}{p{1.8cm} p{0.9cm} p{1.5cm} p{1.5cm} p{1.5cm} p{1.3cm} p{1.3cm} p{1.3cm}}
\caption{Overview of the corpus, presenting a detailed account of target words.}
\label{tab:Overview-of-Corpus}\\
\toprule
\textbf{\begin{tabular}[c]{@{}c@{}}Word\end{tabular}} &
\textbf{\begin{tabular}[c]{@{}c@{}}No. of\\ Senses\end{tabular}} &
\textbf{\begin{tabular}[c]{@{}c@{}}No. of\\ sentences\\ for T1\end{tabular}} &
\textbf{\begin{tabular}[c]{@{}c@{}}No. of\\ sentences\\ for T2\end{tabular}} &
\textbf{\begin{tabular}[c]{@{}c@{}}No. of\\ sentences\\ for T3\end{tabular}} &
\textbf{\begin{tabular}[c]{@{}c@{}}Labels\\ for\\ T1--T2\end{tabular}} &
\textbf{\begin{tabular}[c]{@{}c@{}}Labels\\for\\ T1--T3\end{tabular}} &
\textbf{\begin{tabular}[c]{@{}c@{}}Labels\\for\\ T2--T3\end{tabular}} \\
\midrule
\endfirsthead

\caption*{\textbf{Table \thetable\ (continued): Overview of the corpus.}}\\
\toprule
\textbf{\begin{tabular}[c]{@{}c@{}}Word\end{tabular}} &
\textbf{\begin{tabular}[c]{@{}c@{}}No. of\\ Senses\end{tabular}} &
\textbf{\begin{tabular}[c]{@{}c@{}}No. of\\ sentences\\ for T1\end{tabular}} &
\textbf{\begin{tabular}[c]{@{}c@{}}No. of\\ sentences\\ for T2\end{tabular}} &
\textbf{\begin{tabular}[c]{@{}c@{}}No. of\\ sentences\\ for T3\end{tabular}} &
\textbf{\begin{tabular}[c]{@{}c@{}}Labels\\for\\ T1--T2\end{tabular}} &
\textbf{\begin{tabular}[c]{@{}c@{}}Labels\\for\\ T1--T3\end{tabular}} &
\textbf{\begin{tabular}[c]{@{}c@{}}Labels\\for\\ T2--T3\end{tabular}} \\
\midrule
\endhead

\midrule
\multicolumn{8}{r}{\textit{Continued on next page}}\\
\endfoot

\bottomrule
\endlastfoot

Abc & 3 & 400 & 237 & 1000 & NC & SA & SA \\
Artichoke & 6 & 38 & 39 & 1000 & NC & SA & SA \\
Atm & 4 & 36 & 87 & 1000 & SR & SA & SA \\
Bam & 6 & 84 & 211 & 1000 & SA & SA & SA \\
Battery & 5 & 542 & 347 & 1000 & NC & SA, SR & SA, SR \\
Beetle & 5 & 112 & 120 & 1000 & SA & SA & SA \\
Bing & 9 & 119 & 89 & 1000 & SA & SA & SA \\
Blender & 6 & 112 & 119 & 1000 & SR & SA & SA \\
BMW & 3 & 91 & 141 & 865 & SR & SA & SA \\
Bot & 7 & 143 & 69 & 1000 & NC & SA & SA \\
Bouncer & 5 & 55 & 49 & 1000 & SR & SA & SA \\
Breather & 4 & 46 & 44 & 1000 & SR & SA & SA \\
Brownie & 8 & 73 & 72 & 819 & SA & SA & SA \\
Bush & 7 & 1000 & 1000 & 1000 & SR & SA & SA \\
Cheese & 8 & 1000 & 1000 & 1000 & SA & SA, SR & SA, SR \\
Cheesy & 8 & 52 & 65 & 1000 & SA, SR & SA & SA \\
Chronic & 5 & 482 & 380 & 823 & SA & SA & SA \\
Chump & 4 & 106 & 37 & 1000 & SR & SA & SA \\
Climber & 4 & 75 & 50 & 517 & NC & SA & SA \\
Clip & 8 & 340 & 246 & 1000 & SA, SR & SA & SA \\
Cooker & 6 & 30 & 29 & 1000 & NC & SA & SA \\
Cook & 8 & 1000 & 1000 & 1000 & SA & SA & SA, SR \\
Cool & 6 & 1000 & 1000 & 1000 & NC & SA & SA \\
Cooler & 6 & 397 & 366 & 1000 & SR & SA, SR & SA \\
Coon & 5 & 78 & 38 & 1000 & SR & NC & SA \\
Cosmo & 5 & 61 & 71 & 1000 & NC & SA & SA \\
Crush & 8 & 521 & 399 & 1000 & SA & SA, SR & SA, SR \\
Cucumber & 3 & 74 & 85 & 892 & SA, SR & SA & SA \\
Dinosaur & 6 & 206 & 125 & 1000 & NC & SA & SA \\
Douche & 3 & 36 & 34 & 1000 & SA & SA & SR \\
Drag & 11 & 1000 & 687 & 1000 & SR & SA & SA \\
Eat & 6 & 975 & 995 & 1000 & NC & NC & NC \\
Epic & 2 & 313 & 207 & 1000 & NC & NC & NC \\
Fan & 5 & 1000 & 891 & 1000 & SR & SR & NC \\
Femme & 3 & 38 & 23 & 1000 & SR & NC & SA \\
Fig & 7 & 870 & 230 & 1000 & SR & SA, SR & SA \\
Flutter & 9 & 167 & 145 & 1000 & SA & SA & SA \\
Foam & 5 & 357 & 255 & 1000 & SR & SA & SA \\
Sketch & 6 & 422 & 233 & 1000 & SA, SR & SA, SR & SA, SR \\
Skinny & 7 & 684 & 502 & 1000 & SA, SR & SA, SR & SA, SR \\
Salty & 6 & 214 & 165 & 887 & NC & NC & NC \\
Scum & 3 & 276 & 107 & 1000 & NC & NC & NC \\
Spring & 7 & 100 & 100 & 1000 & SA & SA & SA \\
Frog & 7 & 385 & 225 & 1000 & SR & SA & SA \\
Garnish & 5 & 108 & 219 & 1000 & SR & SA, SR & SA \\
Gash & 5 & 101 & 95 & 1000 & SR & SA & SA \\
Gasoline & 4 & 562 & 240 & 1000 & SA, SR & SA & SA \\
Gay & 6 & 1000 & 1000 & 1000 & SA, SR & SA & SA \\
Germ & 6 & 112 & 70 & 650 & SA, SR & SA & SA \\
Ghost & 9 & 1000 & 798 & 1000 & SA & SA & SA \\
Gook & 5 & 59 & 6 & 1000 & SR & SA & SA \\
Gosh & 2 & 357 & 241 & 1000 & NC & NC & NC \\
Grab & 12 & 1000 & 1000 & 1000 & SR & SR & SA \\
Grade & 3 & 1000 & 921 & 1000 & SA, SR & NC & SA, SR \\
Grit & 9 & 99 & 129 & 1000 & SR & SA & SA \\
Hag & 2 & 140 & 51 & 1000 & NC & NC & NC \\
Hamburger & 7 & 290 & 122 & 1000 & SR & SA, SR & SA \\
Harp & 7 & 280 & 65 & 1000 & SA, SR & SA, SR & SA \\
Hickey & 5 & 49 & 71 & 1000 & SR & SA, SR & SA \\
Hike & 4 & 448 & 374 & 1000 & SR & SA & SA \\
Hive & 6 & 103 & 72 & 1000 & SR & SA & SA \\
Huff & 5 & 121 & 63 & 1000 & SR & SA & SA \\
Hun & 4 & 131 & 60 & 1000 & SA & SA & SA \\
Hut & 3 & 767 & 303 & 1000 & NC & SR & SR \\
Mammy & 4 & 172 & 22 & 876 & SR & SA & SA \\
Moose & 7 & 248 & 159 & 1000 & SA & SA & SA \\
Mosquito & 3 & 187 & 162 & 1000 & SR & NC & SA \\
Mug & 11 & 347 & 326 & 1000 & SA, SR & SA & SA, SR \\
Penguin & 5 & 158 & 53 & 1000 & SR & SA, SR & SA, SR \\
Player & 5 & 1000 & 1000 & 1000 & SA & SA & NC \\
Posse & 6 & 111 & 59 & 1000 & SA & SA & SA \\
Psych & 8 & 87 & 101 & 1000 & SA & SA & SA \\
Putty & 4 & 50 & 33 & 1000 & SR & SA & SA \\
Pitch & 8 & 100 & 100 & 1000 & SA & SA & SA \\
Quaker & 3 & 114 & 55 & 1000 & NC & NC & NC \\
Rad & 5 & 238 & 32 & 1000 & SA & SA & SR \\
Ratchet & 5 & 35 & 34 & 1000 & SA & SA & SA \\
Riff & 11 & 57 & 31 & 1000 & SA & SA & SA, SR \\
Rodent & 2 & 101 & 67 & 905 & NC & NC & NC \\

\end{longtable}
\endgroup

\section{Sense Inventory Construction and Consolidation}
\label{app:bdlsc-sense-inventory}

To address inconsistencies and noise across dictionary sources, we constructed a standardised and empirically grounded sense inventory for each target word. Because slang definitions often vary across lexicographic resources, we developed a multi‐step process that ensures interpretability, reproducibility, and relevance to our time period (1980--2020).

\subsection*{ Dictionary Sources and Initial Sense Collection}
\label{app:bdlsc-dictionary-sources}

We collected candidate senses from the sources described in the \ref{subsec:bdlsc_sense_inventory}. Senses attested from 1980 onward were considered for inclusion, since our corpora begin in 1980. All possible senses that fell within or could plausibly extend into our time span were initially included.

Importantly, we did not exclude a sense solely because its earliest attestation predates 1980. Many lexical meanings originate long before this period, and earlier start dates do not imply that a sense is irrelevant to later usage. A sense was retained if it either:
\begin{itemize}
    \item Showed evidence of continued use into the post-1980 period, or
    \item Could plausibly extend into this period based on its citation history or absence of evidence of disappearance.
\end{itemize}

\subsection*{Verification and Filtering of Urban Dictionary Senses}
\label{app:bdlsc-ud-filtering}

Because Urban Dictionary (UD) is user-generated and therefore highly variable in quality, we applied a set of strict filters to determine which UD-derived slang senses were sufficiently reliable for inclusion. Both criteria below had to be satisfied for a UD sense to be retained.

\paragraph*{Cross-source and corpus confirmation}
A UD-derived sense was retained only if it satisfied two complementary validation criteria. First, the sense had to be independently attested in at least one additional resource included in the sense inventory. This cross-source confirmation ensured that the sense was not a purely idiosyncratic or user-specific invention, but instead reflected a recognised usage beyond the Urban Dictionary platform. Second, the sense was required to appear at least five times in our corpora (COHA or Twitter). This corpus attestation threshold verified that the sense was actively used in real-world contexts, rather than existing solely as a dictionary entry.

\subsection*{Criteria for Consolidating Overlapping or Similar Senses}
\label{app:bdlsc-sense-consolidation}

Because multiple dictionaries often list senses that differ only minimally (for example, general evaluative slang vs. drug‐specific evaluative slang, or a negative insult vs. a prison‐specific insult), we used the following consolidation criteria to avoid inflating the sense inventory with redundant entries:

\paragraph*{Merging criterion}
A pair (or group) of dictionary senses was merged only if 
\emph{all three} of the following were satisfied:
\begin{enumerate}
  \item The core semantic meaning was identical (e.g., 
  ``to annoy/bother'' in two sources). For example, the entries for 
  \textit{salty} describing it as ``irritated'', ``annoyed'', 
  ``resentful'', and ``sour'' all express the same underlying 
  emotional state and were merged into a single sense.
  \item Pragmatic or register differences were minimal (for 
  example, general colloquial vs slightly more specialised but 
  overlapping use). For instance, one source defined the 
  \textit{salty} sense as ``annoyed'' in general informal speech, 
  while another glossed it as ``resentful'' in online slang; both 
  carry the same casual, evaluative tone, so the slight difference 
  in source register did not justify separate entries.
  \item Examples from the different sources were interchangeable in 
  usage (i.e., contexts where one sense could serve in place of the 
  other without loss of meaning). For example, in the sentence 
  ``he was still salty about losing the match'', the ``annoyed'' 
  and ``resentful'' readings can each be substituted without 
  altering the interpretation, confirming that they form one sense.
\end{enumerate}

\paragraph*{Retention of distinct senses}
If \emph{any one} of the following was true, the senses were 
retained as \textbf{distinct} entries:
\begin{itemize}
  \item They expressed a different semantic or pragmatic function. 
  For example, the \textit{salty} sense denoting a temporary 
  emotional state (``irritated'' or ``resentful'') was kept 
  separate from the sense denoting a stable personal characteristic 
  (``tough'', ``aggressive'', or ``hardened''), since a transient 
  feeling and an enduring trait perform different semantic roles.

  \item They occurred in different usage contexts or registers  
  (e.g., for \textit{germ}: prison slang meaning ``cigarette'' 
  vs.\ youth or internet slang meaning ``an unpleasant person'').

  \item They belonged to different social or domain-specific 
  varieties (e.g., for \textit{chronic}: drug-culture slang meaning 
  ``high-quality cannabis'' vs.\ mainstream vernacular meaning 
  ``long-lasting medical condition'').

  \item They had at least five corpus attestations supporting a 
  unique usage pattern separate from other senses.
\end{itemize}

This policy allowed us to maintain fine-grained distinctions (particularly important for slang), while removing redundant or artificially separated senses.

\subsection*{Handling Broad or Non-Lexical Categories}
\label{app:bdlsc-broad-categories}

Some dictionary sources include categories that are extremely broad or not strictly lexical in nature, such as:
\begin{itemize}
    \item “brand/company/song”
    \item “proper name”
    \item “often used as a collocate”
\end{itemize}

These labels do not represent conventional lexical senses. Instead, they describe how the word is used referentially (e.g., as a name) or structurally (e.g., as a recurring collocate) rather than semantically. Such categories can complicate sense inventories and WSD because they do not correspond to stable meanings of the word itself.

However, after analysing our dataset, we found that these non-lexical uses were not isolated anomalies but appeared frequently and consistently for several target words. For example:
\begin{itemize}
    \item \textit{Brownie} and \textit{Germ} were often used as proper names for people, groups, or creative works  
    (e.g., “Brownie performed live last night”, “Germ released a new track”).

    \item \textit{Chronic} and \textit{Salty} frequently appeared as titles of albums, songs, or brands  
    (e.g., “I\textquotesingle{}ve
     had \emph{Chronic} on repeat all week”).

    \item \textit{Brownie} and \textit{Germ} were also commonly used as collocates rather than meaningful lexical items  
    (e.g., “a brownie troop meeting”, “germ warfare research”).  
    In such cases, the word contributes little or no semantic content specific to the target sense.
\end{itemize}

Because these usage patterns occurred across multiple time periods and formed a substantial portion of the real corpus data, we decided not to discard these instances. Removing them would eliminate genuine occurrences and distort the distribution of senses, especially for words where name-based or collocational uses are frequent.

Instead, we retained these cases as a distinct “name/label or collocate” category when supported by sufficient corpus evidence. This approach preserved the full range of naturally attested usage while keeping these non-lexical items clearly separated from actual semantic senses.

\begin{table*}[h]
\centering
\caption{Hyperparameter configurations for GPT-4o prompting strategies, transformer-based models, and machine learning (ML) models used in the study.}
\label{tab:model_hyperparameters}
\footnotesize

\begin{tabular}{@{}p{0.25\linewidth} p{0.7\linewidth}@{}}  
\toprule

\multicolumn{2}{c}{\textbf{\textbf{LLM Models}}}\\
\midrule
\textbf{Model} & \textbf{parameters} \\
\midrule
\begin{tabular}[t]{@{}l@{}}GPT-4o-mini zero-shot,\\ GPT-4o zero-shot,\\ GPT-4o-mini few-shot,\\ GPT-4o few-shot\\  GPT-3.5-turbo\end{tabular}
& Max Tokens = 2048, Temperature = 0, Top\_p = 0.9, Frequency Penalty = 0.0, Presence Penalty = 0.0 \\

\midrule
\multicolumn{2}{c}{\textbf{Transformer Models}}\\
\midrule
\textbf{Model} & \textbf{Hyperparameters} \\
\midrule
\begin{tabular}[t]{@{}c@{}}Fine-tuned \\Transformer Models\\(BERT, RoBERTa)\end{tabular} &
\texttt{num\_train\_epochs = 30, learning\_rate = 4e-5, train\_batch\_size = 64, eval\_batch\_size = 64, optimiser = AdamW} \\

\midrule
\multicolumn{2}{c}{\textbf{Machine Learning Models}}\\
\midrule
\textbf{Estimator} & \textbf{Hyperparameters} \\
\midrule
Logistic Regression (LR) &
\texttt{penalty = l2, C = 1.0, solver = lbfgs, max\_iter = 100, verbose = 0} \\

Support Vector Machine (SVM) &
\texttt{C = 1.0, gamma = 1.0, cache\_size = 200, max\_iter = -1} \\

Random Forest (RF) &
\texttt{n\_estimators = 100, max\_depth = 10, min\_samples\_split = 2} \\

AdaBoost &
\texttt{n\_estimators = 50, learning\_rate = 1.0, base\_estimator = DecisionTreeClassifier, algorithm = SAMME.R} \\

CatBoost &
\texttt{iterations = 1000, learning\_rate = 0.03, depth = 6, verbose = True} \\
\bottomrule
\end{tabular}
\end{table*}

\begin{sidewaystable*}
\scriptsize
\caption{WSD results across different methods (T1--T3). Highest results are highlighted in gray.}

\label{tab:wsd_results_full}
\setlength{\tabcolsep}{3pt}
\renewcommand{\arraystretch}{1.5}

\resizebox{\textwidth}{!}{%
\begin{tabular}{
@{}l l
p{1.7cm}p{1.6cm}p{0.8cm}p{0.8cm}p{0.8cm}l
p{0.8cm}p{0.8cm}p{0.8cm}l
p{0.8cm}p{0.8cm}p{0.8cm}l@{}}

\toprule
\multirow{2}{*}{\textbf{Word}} & \multirow{2}{*}{\textbf{TP}} &
\multicolumn{6}{c}{\textbf{N-grams and Char N-grams -Based ML}} &
\multicolumn{4}{c}{\textbf{DistilBERT-Based ML}} &
\multicolumn{4}{c}{\textbf{FastText-Based ML}}\\
\addlinespace[1pt]
 & & Feat. & Model & Acc & Mac-F1 & Mic-F1 & Pred.\ Senses &
 Acc & Mac-F1 & Mic-F1 & Pred.\ Senses &
 Acc & Mac-F1 & Mic-F1 & Pred.\ Senses\\
\midrule

\multirow{3}{*}{Eat} 
& T1 & Uni-gram & AdaBoost & 0.57 & 0.17 & 0.60 & S1, S2, S3, S5 &
0.13 & 0.10 & 0.16 & S1, S3 &
0.26 & 0.18 & 0.32 & S1, S2, S5\\

& T2 & Uni-gram & AdaBoost & 0.60 & 0.17 & 0.65 & S1, S2, S3 &
0.14 & 0.09 & 0.20 & S1, S2, S3 &
0.29 & 0.25 & 0.32 & S1, S4, S5\\

& T3 & Uni-gram & AdaBoost & 0.79 & 0.20 & 0.75 & S1, S2, S3, S4 &
0.41 & 0.20 & 0.52 & S1, S3 &
0.38 & 0.17 & 0.49 & S1, S2, S5\\

\midrule

\multirow{3}{*}{BMW} 
& T1 & Bi-gram & AdaBoost & 0.91 & 0.45 & 0.93 & S1 &
0.18 & 0.10 & 0.30 & S1, S2 &
0.20 & 0.12 & 0.17 & S1, S2 \\

& T2 & Bi-gram & AdaBoost & 0.97 & 0.33 & 0.99 & S1 &
0.17 & 0.10 & 0.29 & S1 &
0.37 & 0.17 & 0.22 & S1 \\

& T3 & Bi-gram & AdaBoost & 0.98 & 0.40 & 0.98 & S1, S2 &
0.67 & 0.28 & 0.79 & S1, S2, S3 &
0.61 & 0.28 & 0.75 & S1, S2, S3 \\

\midrule

\multirow{3}{*}{Brownie} 
& T1 & Bi-gram & SVM & 0.49 & 0.21 & 0.42 & S1, S3, S6 &
0.16 & 0.10 & 0.24 & S1, S2, S3 &
0.20 & 0.13 & 0.10 & S1, S3 \\

& T2 & Comb.\ 1–2 gram & AdaBoost & 0.58 & 0.22 & 0.58 & S1, S2, S3, S4, S6, S7 &
0.20 & 0.10 & 0.23 & S1, S3, S7 &
0.15 & 0.17 & 0.22 & S1, S3, S6, S7 \\

& T3 & Comb.\ 1–2 gram & AdaBoost & 0.73 & 0.22 & 0.63 & S1, S6 &
0.32 & 0.18 & 0.40 & S1, S2, S3, S4, S5, S6 &
0.26 & 0.18 & 0.32 & S1, S2, S7, S8 \\

\midrule

\multirow{3}{*}{Chronic}
& T1 & Comb.\ word 1–3 gram & SVM & 0.83 & 0.23 & 0.79 & S2, S4 &
0.18 & 0.09 & 0.29 & S1, S2 &
0.44 & 0.19 & 0.57 & S1, S2 \\

& T2 & Uni-gram & SVM &\cellcolor{gray!20} 0.87 & \cellcolor{gray!20}0.21 &\cellcolor{gray!20} 0.83 & \cellcolor{gray!20}S2, S4 &
0.18 & 0.10 & 0.28 & S1, S2, S3 &
0.38 & 0.17 & 0.51 & S1, S2 \\

& T3 & Comb.\ Char 1–3 gram & CatBoost & 0.58 & 0.36 & 0.65 & S1, S2, S3, S4, S5 &
0.29 & 0.23 & 0.36 & S1, S2, S4 &
0.48 & 0.31 & 0.55 & S1, S2, S4 \\

\midrule

\multirow{3}{*}{Climber}
& T1 & Comb. word 1–3 gram & SVM & 0.73 & 0.57 & 0.78 & S1, S2, S3 &
0.57 & 0.51 & 0.63 & S1, S2 &
0.48 & 0.41 & 0.50 & S1, S3 \\

& T2 & Comb. word 1–3 gram & AdaBoost & 0.74 & 0.42 & 0.77 & S1, S2, S3 &
0.64 & 0.51 & 0.73 & S1, S2 &
0.46 & 0.36 & 0.55 & S1, S2 \\

& T3 & Comb. word 1–3 gram & AdaBoost & 0.85 & 0.48 & 0.85 & S1, S2, S3, S4 &
0.64 & 0.40 & 0.69 & S1, S2, S3  &
0.66 & 0.42 & 0.71 & S1, S2 \\

\bottomrule
\end{tabular}%
}
\end{sidewaystable*}

\begin{sidewaystable*}
\scriptsize
\caption*{Table~\ref{tab:wsd_results_full} (Continued) WSD results across different methods (T1--T3). Highest results are highlighted in gray.}

\label{tab:wsd_results_part2}
\setlength{\tabcolsep}{3pt}
\renewcommand{\arraystretch}{1.5}

\resizebox{\textwidth}{!}{%
\begin{tabular}{
@{}l l
p{1.7cm}p{1.6cm}p{0.8cm}p{0.8cm}p{0.8cm}l
p{0.8cm}p{0.8cm}p{0.8cm}l
p{0.8cm}p{0.8cm}p{0.8cm}l@{}}

\toprule
\multirow{2}{*}{\textbf{Word}} & \multirow{2}{*}{\textbf{TP}} &
\multicolumn{6}{c}{\textbf{N-grams and Char N-grams}} &
\multicolumn{4}{c}{\textbf{DistilBERT}} &
\multicolumn{4}{c}{\textbf{FastText}}\\
\addlinespace[1pt]
 & & \textbf{Feat.} & \textbf{Model} & \textbf{Acc} & \textbf{Mac-F1} & \textbf{Mic-F1} & \textbf{Pred.\ Senses} &
   \textbf{Acc} & \textbf{Mac-F1} & \textbf{Mic-F1} & \textbf{Pred.\ Senses} &
   \textbf{Acc} & \textbf{Mac-F1} & \textbf{Mic-F1} & \textbf{Pred.\ Senses}\\
\midrule

\multirow{3}{*}{Cucumber}
& T1 & Uni-gram & SVM & \cellcolor{gray!20}0.97 & \cellcolor{gray!20}0.33 & \cellcolor{gray!20}0.98 & \cellcolor{gray!20}S1, S2 &
0.31 & 0.17 & 0.46 & S1, S2 &
0.50 & 0.24 & 0.65 & S1, S2 \\
& T2 & Uni-gram & SVM & \cellcolor{gray!20}0.92 & \cellcolor{gray!20}0.45 & \cellcolor{gray!20}0.90 & \cellcolor{gray!20}S2, S3 &
0.67 & 0.32 & 0.75 & S1, S2 &
0.76 & 0.39 & 0.82 & S2, S3 \\
& T3 & Bi-gram & AdaBoost & 0.83 & 0.30 & 0.75 & S1, S2, S3 &
0.63 & 0.47 & 0.69 & S1, S2, S3 &
0.74 & 0.46 & 0.75 & S1, S2, S3 \\
\midrule

\multirow{3}{*}{Germ}
& T1 & Comb. word 1–3 gram & LR & 0.48 & 0.28 & 0.47 & S1, S3, S4, S5, S6 &
0.47 & 0.21 & 0.47 & S1, S3, S6 &
0.26 & 0.16 & 0.28 & S1, S3, S6 \\
& T2 & Comb. word 1–3 gram & LR & 0.50 & 0.39 & 0.45 & S1, S3, S5, S6 &
0.43 & 0.23 & 0.40 & S1, S3, S6 &
0.37 & 0.21 & 0.36 & S1, S3, S6 \\
& T3 & Comb. word 1–3 gram & SVM & 0.39 & 0.31 & 0.39 & S1, S3, S4, S5, S6 &
0.34 & 0.31 & 0.35 & S1, S2, S3, S4, S5, S6 &
0.30 & 0.25 & 0.32 & S1, S2, S5, S6 \\
\midrule

\multirow{3}{*}{Mammy}
& T1 & Comb. word 1–3 gram & RFC & 0.48 & 0.23 & 0.42 & S1, S2, S4 &
0.41 & 0.23 & 0.36 & S1, S2, S4 &
0.43 & 0.28 & 0.48 & S1, S2 \\
& T2 & Char Bi-gram & RFC & 0.50 & 0.22 & 0.46 & S1, S2 &
0.59 & 0.35 & 0.68 & S1, S2 &
0.36 & 0.24 & 0.48 & S1, S2 \\
& T3 & Comb. word 1–3 gram & CatBoost & 0.72 & 0.26 & 0.75 & S1, S2, S3, S4 &
0.51 & 0.22 & 0.61 & S1, S2, S4 &
0.44 & 0.26 & 0.47 & S1 \\
\midrule

\multirow{3}{*}{Rodent}
& T1 & Bi-gram & SVM & 0.74 & 0.54 & 0.67 & S1, S2 &
0.76 & 0.66 & 0.74 & S1, S2 &
0.70 & 0.67 & 0.71 & S1, S2 \\
& T2 & Uni-gram & CatBoost & 0.66 & 0.47 & 0.56 & S1, S2 &
0.68 & 0.62 & 0.66 & S1, S2 &
0.60 & 0.59 & 0.61 & S1, S2 \\
& T3 & Bi-gram & SVM & 0.68 & 0.47 & 0.59 & S1, S2 &
0.63 & 0.58 & 0.63 & S1, S2 &
0.66 & 0.62 & 0.66 & S1, S2 \\
\midrule

\multirow{3}{*}{Salty}
& T1 & Uni-gram & SVM & 0.41 & 0.20 & 0.51 & S1, S2, S3, S4, S5 &
0.15 & 0.14 & 0.13 & S2, S3, S4, S5 &
0.15 & 0.12 & 0.20 & S1, S4, S5 \\
& T2 & Uni-gram & SVM & 0.43 & 0.19 & 0.53 & S1, S2, S3, S5, S6 &
0.17 & 0.12 & 0.15 & S2, S3, S4 &
0.14 & 0.10 & 0.15 & S1, S2 \\
& T3 & Uni-gram & RFC & 0.23 & 0.16 & 0.30 & S1, S2, S3, S6 &
0.21 & 0.15 & 0.30 & S1, S2, S3, S5 &
0.18 & 0.14 & 0.24 & S1, S2, S3 \\
\bottomrule
\end{tabular}%
}
\end{sidewaystable*}


\begin{sidewaystable*}[h]
\small
\caption*{Table~\ref{tab:wsd_results_full} (Continued) WSD results across different methods (T1--T3). Highest results are highlighted in gray.}
\label{tab:wsd_results_part3}
\setlength{\tabcolsep}{3pt}
\renewcommand{\arraystretch}{2.2}

\resizebox{\textwidth}{!}{
\begin{tabular}{
@{}l l
cccc cccc  cccc  cccc@{}}
\toprule
\multirow{2}{*}{\textbf{Word}} &
\multirow{2}{*}{\textbf{TP}} &
\multicolumn{4}{c}{\textbf{GPT-4o-mini Zero-shot}} &
\multicolumn{4}{c}{\textbf{GPT-4o Zero-shot}} &
\multicolumn{4}{c}{\textbf{GPT-4o-mini Few-shot}} &
\multicolumn{4}{c}{\textbf{GPT-4o Few-shot}} \\
 & &
\textbf{Acc} &
\shortstack{\textbf{Macro-}\\\textbf{F1}} &
\shortstack{\textbf{Micro-}\\\textbf{F1}} &
\shortstack{\textbf{Pred.}\\\textbf{Senses}} &
\textbf{Acc} &
\shortstack{\textbf{Macro-}\\\textbf{F1}} &
\shortstack{\textbf{Micro-}\\\textbf{F1}} &
\shortstack{\textbf{Pred.}\\\textbf{Senses}} &
\textbf{Acc} &
\shortstack{\textbf{Macro-}\\\textbf{F1}} &
\shortstack{\textbf{Micro-}\\\textbf{F1}} &
\shortstack{\textbf{Pred.}\\\textbf{Senses}} &
\textbf{Acc} &
\shortstack{\textbf{Macro-}\\\textbf{F1}} &
\shortstack{\textbf{Micro-}\\\textbf{F1}} &
\shortstack{\textbf{Pred.}\\\textbf{Senses}} \\
\midrule

\multirow{3}{*}{Eat}
& T1 & 0.93 & 0.43 & 0.93 & S1, S2, S3, S5, S6 &
0.89 & 0.41 & 0.89 & S1, S2, S3, S5, S6 &
\cellcolor{gray!20}0.95 & \cellcolor{gray!20}0.58 & \cellcolor{gray!20}0.95 & \cellcolor{gray!20}S1, S2, S3, S5, S6 &
0.94 & 0.52 & 0.94 & S1, S2, S3, S5, S6 \\
& T2 & 0.94 & 0.40 & 0.94 & S1, S2, S3, S5, S6 &
\cellcolor{gray!20}0.96 & \cellcolor{gray!20}0.44 & \cellcolor{gray!20}0.96 & \cellcolor{gray!20}S1, S2, S5 &
0.94 & 0.53 & 0.94 & S1, S2, S3, S5, S6 &
0.78 & 0.54 & 0.78& S1, S2, S3, S5, S6 \\
& T3 & 0.93 & 0.41 & 0.93 & S1, S2, S3, S5, S6 &
0.95 & 0.46 & 0.95 & S1, S2, S3, S5, S6 &
0.95 & 0.57 & 0.95 & S1, S2, S3, S6 &
\cellcolor{gray!20}0.96 & \cellcolor{gray!20}0.59 & \cellcolor{gray!20}0.96 & \cellcolor{gray!20}S1, S2, S3, S5, S6 \\
\midrule

\multirow{3}{*}{BMW}
& T1 & 0.96 & 0.44 & 0.96 & S1 &
0.96 & 0.66 & 0.96 & S1 &
\cellcolor{gray!20}0.98 & \cellcolor{gray!20}0.83 & \cellcolor{gray!20}0.98 & \cellcolor{gray!20}S1 &
0.96 & 0.66 & 0.96 & S1 \\
& T2 & \cellcolor{gray!20}1.00 & \cellcolor{gray!20}1.00 & \cellcolor{gray!20}1.00 & \cellcolor{gray!20}S1 &
0.97 & 0.46 & 0.97 & S1 &
\cellcolor{gray!20}1.00 & \cellcolor{gray!20}1.00 & \cellcolor{gray!20}1.00 & \cellcolor{gray!20}S1 &
0.99 & 0.50 & 0.99 & S1 \\
& T3 & 0.99 & 0.68 & 0.99 & S1, S2 &
0.97 & 0.47 & 0.97 & S1, S2 &
\cellcolor{gray!20}0.99 & \cellcolor{gray!20}0.71 & \cellcolor{gray!20}0.99 & \cellcolor{gray!20}S1, S2 &
0.97 & 0.46 & 0.97 & S1, S2 \\
\midrule

\multirow{3}{*}{Brownie}
& T1 & 0.49 & 0.21 & 0.42 & S1, S4, S6, S7, S8 &
0.57 & 0.10 & 0.57 & S1, S4, S6, S7, S8 &
0.20 & 0.13 & 0.10 & S1, S5, S6, S7, S8 &
0.45 & 0.40 & 0.45 & S1, S2, S6, S7, S8 \\

& T2 & 0.58 & 0.22 & 0.58 & S1, S2, S4, S6, S7, S8 &
0.52 & 0.10 & 0.52 & S1, S3, S4, S6, S7, S8 &
0.15 & 0.17 & 0.22 & S1, S3, S4, S6, S7, S8 &
\cellcolor{gray!20}0.76 & \cellcolor{gray!20}0.54 & \cellcolor{gray!20}0.76 & \cellcolor{gray!20}S1, S3, S4, S6, S7, S8 \\

& T3 & 0.73 & 0.22 & 0.63 & S1, S2, S3, S4, S6, S8 &
0.69& 0.38 & 0.69 & S1, S2, S3, S4, S6, S7, S8 &
0.69 & 0.38 &0.69 & S1, S2, S3, S4, S5, S6, S7, S8 &
\cellcolor{gray!20}0.80 & \cellcolor{gray!20}0.46 & \cellcolor{gray!20}0.80 & \cellcolor{gray!20}S1, S2, S3, S4, S5, S6, S7, S8 \\
\midrule

\multirow{3}{*}{Chronic}
& T1 & 0.81 & 0.41 & 0.81 & S1, S2 &
0.79 & 0.36 & 0.79 & S1, S2 &
\cellcolor{gray!20}0.86 & \cellcolor{gray!20}0.45 & \cellcolor{gray!20}0.86 & \cellcolor{gray!20}S1, S2 &
0.84 & 0.43 & 0.84 & S1, S2 \\

& T2 & 0.83 & 0.38 & 0.83 & S1, S2, S4 &
\cellcolor{gray!20}0.87 & \cellcolor{gray!20}0.43 & \cellcolor{gray!20}0.87 & \cellcolor{gray!20}S1, S2, S4 &
0.84 & 0.40 & 0.84 & S1, S2, S4&
0.85 & 0.41 & 0.85 & S1, S2, S3, S4\\

& T3 & 0.80 & 0.36 & 0.80 & S1, S2, S3, S4 , S5 &
0.83 & 0.39 & 0.83 & S1, S2, S3, S4, S5  &
0.84 & 0.41 & 0.84 & S1, S2, S3, S4 , S5  &
\cellcolor{gray!20}0.88 & \cellcolor{gray!20}0.46 & \cellcolor{gray!20}0.88 & \cellcolor{gray!20}S1, S2, S3, S4, S5  \\
\midrule

\multirow{3}{*}{Climber}
& T1 & \cellcolor{gray!20}0.95 & \cellcolor{gray!20}0.94 & \cellcolor{gray!20}0.95 & \cellcolor{gray!20}S1, S2 &
0.93 & 0.91 & 0.93 & S1, S2, S3 &
0.88 & 0.66 & 0.88 & S1, S2&
0.93 & 0.70 & 0.93 & S1, S2, S3 \\

& T2 & \cellcolor{gray!20}0.94 &\cellcolor{gray!20} 0.88 &\cellcolor{gray!20} 0.94 &\cellcolor{gray!20} S1, S2 &
0.91 & 0.28 & 0.91 & S1, S2 &
0.90 & 0.83 & 0.90 & S1, S2 &
0.80 & 0.90 & 0.80& S1, S2, S3 \\

& T3 & \cellcolor{gray!20}0.91 & \cellcolor{gray!20}0.57 & \cellcolor{gray!20}0.91 & \cellcolor{gray!20}S1, S2, S3 &
\cellcolor{gray!20}0.91 & \cellcolor{gray!20}0.28 &\cellcolor{gray!20} 0.91 &\cellcolor{gray!20} S1, S2, S3 &
0.80 & 0.47 & 0.80 & S1, S2, S3 &
\cellcolor{gray!20}0.91 & \cellcolor{gray!20}0.28 & \cellcolor{gray!20}0.91 & \cellcolor{gray!20}S1, S2, S3 \\
\bottomrule
\end{tabular}}
\end{sidewaystable*}

\begin{sidewaystable*}[h]
\small
\caption*{Table~\ref{tab:wsd_results_full} (Continued) WSD results across different methods (T1--T3). Highest results are highlighted in gray.}
\label{tab:wsd_results_part4}
\setlength{\tabcolsep}{3pt}
\renewcommand{\arraystretch}{2.0}
\resizebox{\textwidth}{!}{
\begin{tabular}{
@{}l l
cccc cccc cccc cccc@{}}
\toprule
\multirow{2}{*}{\textbf{Word}} &
\multirow{2}{*}{\textbf{TP}} &
\multicolumn{4}{c}{\textbf{GPT-4o-mini Zero-shot}} &
\multicolumn{4}{c}{\textbf{GPT-4o Zero-shot}} &
\multicolumn{4}{c}{\textbf{GPT-4o-mini Few-shot}} &
\multicolumn{4}{c}{\textbf{GPT-4o Few-shot}} \\
 & &
\textbf{Acc} &
\shortstack{\textbf{Macro-}\\\textbf{F1}} &
\shortstack{\textbf{Micro-}\\\textbf{F1}} &
\shortstack{\textbf{Pred.}\\\textbf{Senses}} &
\textbf{Acc} &
\shortstack{\textbf{Macro-}\\\textbf{F1}} &
\shortstack{\textbf{Micro-}\\\textbf{F1}} &
\shortstack{\textbf{Pred.}\\\textbf{Senses}} &
\textbf{Acc} &
\shortstack{\textbf{Macro-}\\\textbf{F1}} &
\shortstack{\textbf{Micro-}\\\textbf{F1}} &
\shortstack{\textbf{Pred.}\\\textbf{Senses}} &
\textbf{Acc} &
\shortstack{\textbf{Macro-}\\\textbf{F1}} &
\shortstack{\textbf{Micro-}\\\textbf{F1}} &
\shortstack{\textbf{Pred.}\\\textbf{Senses}} \\
\midrule

\multirow{3}{*}{Cucumber}
& T1 & 0.95 & 0.49 & 0.95 &  S1, S2, S3 &
0.65 & 0.44 & 0.65 & S1, S2 &
0.88 & 0.39 & 0.88 & S1, S2, S3 &
0.95 & 0.32 & 0.95 & S1, S2 \\

& T2 & 0.87 & 0.31 & 0.87 & S1, S2, S3 &
0.80 & 0.32 & 0.80 & S1, S2, S3 &
0.88 & 0.31 & 0.88 & S1, S2, S3 &
0.89 & 0.31 & 0.89 & S1, S2, S3 \\

& T3 & 0.88 & 0.72 & 0.88 & S1, S2, S3 &
\cellcolor{gray!20}0.91 & \cellcolor{gray!20}0.33 & \cellcolor{gray!20}0.91 & \cellcolor{gray!20}S1, S2, S3 &
0.88 & 0.76 & 0.88 & S1, S2, S3 &
0.88 & 0.55 & 0.88 & S1, S2, S3 \\
\midrule

\multirow{3}{*}{Germ}
& T1 & 0.31 & 0.17 & 0.31 & S1, S3, S4, S6&
0.53 & 0.31 & 0.53 & S1, S3, S4, S6 &
0.49 & 0.29 & 0.49 & S1, S4, S5, S6 &
0.54 & 0.32 & 0.54 & S1, S3, S4, S6 \\

& T2 & 0.43 & 0.27 & 0.43 & S1, S3, S4, S5, S6 &
0.44 & 0.51 & 0.44 & S1, S3, S5, S6 &
0.44 & 0.32 & 0.44 & S1, S3, S4, S5, S6 &
0.55 & 0.32 & 0.55 & S1, S3, S4, S5, S6 \\

& T3 & 0.44 & 0.33 & 0.44 & S1, S2, S3, S4, S5, S6 &
0.44 & 0.51 & 0.44 & S1, S2, S3, S4, S5, S6 &
0.50 & 0.41 & 0.50 & S1, S2, S3, S4, S5, S6 &
0.44 & 0.51 & 0.44 & S1, S2, S3, S4, S5, S6\\
\midrule

\multirow{3}{*}{Mammy}
& T1 & 0.55 & 0.42 & 0.55 & S1, S2, S3, S4 &
\cellcolor{gray!20}0.65 & \cellcolor{gray!20}0.44 & \cellcolor{gray!20}0.65 & \cellcolor{gray!20}S1, S2, S4 &
0.60 & 0.52 & 0.60 & S1, S2, S4 &
0.59 & 0.42 & 0.59 & S1, S2, S4 \\

& T2 & 0.70 & 0.55 & 0.70 & S1, S2, S4 &
0.70 & 0.55 & 0.70 & S1, S2, S4 &
\cellcolor{gray!20}0.81 & \cellcolor{gray!20}0.28 & \cellcolor{gray!20}0.81 & \cellcolor{gray!20}S1, S2, S4 &
0.69 & 0.62 & 0.69 & S1, S2\\

& T3 & 0.72 & 0.47 & 0.72 & S1, S2, S3, S4 &
0.78 & 0.43 & 0.78 & S1, S2, S3, S4 &
0.72 & 0.52 & 0.72 & S1, S2, S3, S4 &
\cellcolor{gray!20}0.80 & \cellcolor{gray!20}0.58 & \cellcolor{gray!20}0.80 & \cellcolor{gray!20}S1, S2, S3, S4 \\
\midrule

\multirow{3}{*}{Rodent}
& T1 & \cellcolor{gray!20}0.95 & \cellcolor{gray!20}0.49 & \cellcolor{gray!20}0.95 & \cellcolor{gray!20}S1, S2 &
0.88 & 0.82 & 0.88 & S1, S2 &
0.84 & 0.79 & 0.84 & S1, S2 &
0.83 & 0.77 & 0.83 & S1, S2 \\
& T2 & 0.72 & 0.67 & 0.72 & S1, S2 &
\cellcolor{gray!20}0.91 & \cellcolor{gray!20}0.28 & \cellcolor{gray!20}0.91 & \cellcolor{gray!20}S1, S2 &
0.84 & 0.80 & 0.84 & S1, S2 &
0.84 & 0.80 & 0.84 & S1, S2 \\
& T3 & 0.88 & 0.86 & 0.88 & S1, S2 &
\cellcolor{gray!20}0.91 & \cellcolor{gray!20}0.28 & \cellcolor{gray!20}0.91 & \cellcolor{gray!20}S1, S2 &
0.87 & 0.86 & 0.87 & S1, S2 &
0.89 & 0.35 & 0.89 & S1, S2 \\
\midrule

\multirow{3}{*}{Salty}
& T1 & 0.65 & 0.38 & 0.65 & S1, S2, S3, S4, S6 &
\cellcolor{gray!20}0.75 & \cellcolor{gray!20}0.40 & \cellcolor{gray!20}0.75 & \cellcolor{gray!20}S1, S2, S3, S4, S5, S6 &
0.74 & 0.38 & 0.74 & S1, S2, S3, S4, S5, S6 &
0.70 & 0.40 & 0.70 & S1, S2, S3, S4, S5, S6 \\

& T2 & 0.72 & 0.40 & 0.72 & S1, S2, S4, S5, S6  &
0.72 & 0.40 & 0.72 & S1, S2, S4, S5, S6  &
0.75 & 0.46 & 0.75 & S1, S2, S4, S5, S6  &
\cellcolor{gray!20}0.84 & \cellcolor{gray!20}0.58 & \cellcolor{gray!20}0.84 & \cellcolor{gray!20}S1, S2, S4, S5, S6  \\

& T3 & 0.78 & 0.40 & 0.78 & S1, S2, S3, S4, S6 &
0.81 & 0.46 & 0.81 & S1, S2, S3, S4, S6 &
0.79 & 0.50 & 0.79 & S1, S2, S3, S4, S6 &
\cellcolor{gray!20}0.85 & \cellcolor{gray!20}0.55 & \cellcolor{gray!20}0.85 & \cellcolor{gray!20}S1, S2, S3, S4, S6 \\
\bottomrule
\end{tabular}}
\end{sidewaystable*}

\begin{sidewaystable*}[h]
\small
\caption*{Table~\ref{tab:wsd_results_full} (Continued) WSD results across different methods (T1--T3). Highest results are highlighted in gray.}
\label{tab:wsd_results_part5}
\setlength{\tabcolsep}{3pt}
\renewcommand{\arraystretch}{1.9}
\resizebox{\textwidth}{!}{
\begin{tabular}{
@{}c l cccc cccc c l cccc cccc@{}}
\toprule
\multirow{2}{*}{\textbf{Word}} &
\multirow{2}{*}{\textbf{TP}} &
\multicolumn{4}{c}{\textbf{BERT-large}} &
\multicolumn{4}{c }{\textbf{RoBERTa-large}} &
\multirow{2}{*}{\textbf{Word}} &
\multirow{2}{*}{\textbf{TP}} &
\multicolumn{4}{c }{\textbf{BERT-large}} &
\multicolumn{4}{c}{\textbf{RoBERTa-large}} \\
&&
\textbf{Acc} &
\shortstack{\textbf{Macro-}\\\textbf{F1}} &
\shortstack{\textbf{Micro-}\\\textbf{F1}} &
\shortstack{\textbf{Pred.}\\\textbf{Senses}} &
\textbf{Acc} &
\shortstack{\textbf{Macro-}\\\textbf{F1}} &
\shortstack{\textbf{Micro-}\\\textbf{F1}} &
\shortstack{\textbf{Pred.}\\\textbf{Senses}} &&&
\textbf{Acc} &
\shortstack{\textbf{Macro-}\\\textbf{F1}} &
\shortstack{\textbf{Micro-}\\\textbf{F1}} &
\shortstack{\textbf{Pred.}\\\textbf{Senses}} &
\textbf{Acc} &
\shortstack{\textbf{Macro-}\\\textbf{F1}} &
\shortstack{\textbf{Micro-}\\\textbf{F1}} &
\shortstack{\textbf{Pred.}\\\textbf{Senses}} \\
\midrule

\multirow{3}{*}{Eat}
& T1 & 0.80 & 0.50 & 0.80 & S1, S2, S3, S5, S6 &
0.80 & 0.47 & 0.80 & S1, S2, S3, S6 &
\multirow{3}{*}{Cucumber}
& T1 & 0.67 & 0.41 & 0.67 & S1, S2 &
0.65 & 0.47 & 0.65 & S2 \\

& T2 & 0.86 & 0.52 & 0.86 & S1, S2, S5 &
0.72 & 0.52 & 0.72 & S1, S2 &
& T2 & 0.68 & 0.42 & 0.68 & \shortstack{S1, S2, S3} &
0.68 & 0.52 & 0.68 & \shortstack{S1, S2, S3} \\

& T3 & 0.83 & 0.48 & 0.83 & S1, S2, S3, S5, S6 &
0.73 & 0.63 & 0.73 & S1, S2, S3, S5, S6 &
& T3 & 0.70 & 0.44 & 0.70 & \shortstack{S1, S2, S3} &
0.76 & 0.63 & 0.76 & \shortstack{S1, S2, S3} \\
\midrule

\multirow{3}{*}{BMW}
& T1 & 0.86 & 0.60 & 0.86 & S1 &
0.75 & 0.47 & 0.75 & S1 &
\multirow{3}{*}{Germ}
& T1 & 0.60 & 0.37 & 0.60 & \shortstack{S1, S3, S4, S6} &
\cellcolor{gray!20}0.62 &\cellcolor{gray!20} 0.47 & \cellcolor{gray!20}0.69 & \cellcolor{gray!20}\shortstack{S1, S3, S4, S6} \\

& T2 & 0.83 & 0.61 & 0.83 & S1 &
0.74 & 0.52 & 0.74 & S1 &
& T2 & 0.63 & 0.40 & 0.63 & \shortstack{S1, S3, S5, S6} &
\cellcolor{gray!20}0.68 & \cellcolor{gray!20}0.52 &\cellcolor{gray!20} 0.68 &\cellcolor{gray!20} \shortstack{S1, S3, S5} \\

& T3 & 0.85 & 0.64 & 0.85 & \shortstack{S1, S2} &
0.80 & 0.63 & 0.80 & \shortstack{S1, S2} &
& T3 & 0.65 & 0.41 & 0.65 & \shortstack{S1, S2, S3, S4, S5, S6} &
\cellcolor{gray!20}0.78 & \cellcolor{gray!20}0.63 & \cellcolor{gray!20}0.78 & \cellcolor{gray!20}\shortstack{S1, S2, S3, S4, S5, S6} \\
\midrule

\multirow{3}{*}{Brownie}
& T1 & 0.55 & 0.28 & 0.55 & \shortstack{S1, S4, S6, S7, S8} &
\cellcolor{gray!20}0.68 & \cellcolor{gray!20}0.47 & \cellcolor{gray!20}0.68 & \cellcolor{gray!20}\shortstack{S1, S3, S8} &
\multirow{3}{*}{Mammy}
& T1 & 0.60 & 0.40 & 0.60 & \shortstack{S1, S2, S4} &
0.60 & 0.47 & 0.60 & \shortstack{S1, S2, S3, S4}
\\
& T2 & 0.68 & 0.36 & 0.68 & \shortstack{S1, S3, S4, S6, S7, S8} &
0.70 & 0.52 & 0.70 & \shortstack{S1, S2, S3, S4, S6, S7} &
& T2 & 0.69 & 0.44 & 0.69 & \shortstack{S1, S2, S4} &
0.66 & 0.52 & 0.66 & \shortstack{S1, S2, S3} \\
& T3 & 0.74 & 0.40 & 0.74 & \shortstack{S1, S2, S3, S4, S6, S7, S8} &
0.74 & 0.63 & 0.74 & \shortstack{S1, S2, S3, S4, S7, S8} &
& T3 & 0.74 & 0.48 & 0.74 & \shortstack{S1, S2, S3, S4} &
0.68 & 0.63 & 0.68 & \shortstack{S1, S2, S3, S4} \\
\midrule

\multirow{3}{*}{Chronic}
& T1 & 0.70 & 0.40 & 0.70 & S1, S2 &
0.70 & 0.47 & 0.70 & S1, S2 &
\multirow{3}{*}{Rodent}
& T1 & 0.90 & 0.46 & 0.90 & S1, S2 &
0.83 & 0.47 & 0.83 & S1, S2 \\
& T2 & 0.77 & 0.43 & 0.77 & S1, S2, S4 &
0.76 & 0.52 & 0.76 & S1, S2 &
& T2 & 0.84 & 0.45 & 0.84 & S1, S2 &
0.78 & 0.52 & 0.78 & S1, S2 \\

& T3 & 0.77 & 0.42 & 0.77 & S1, S2, S3, S4, S5 &
0.70 & 0.63 & 0.70 & S1, S2, S3, S4, S5 &
& T3 & 0.85 & 0.45 & 0.85 & S1, S2 &
0.87 & 0.63 & 0.87 & S1, S2 \\
\midrule

\multirow{3}{*}{Climber}
& T1 & 0.90 & 0.58 & 0.90 & \shortstack{S1, S2, S3} &
0.72 & 0.47 & 0.72& \shortstack{S1, S2} &
\multirow{3}{*}{Salty}
& T1 & 0.63 & 0.38 & 0.63 & \shortstack{S1, S2, S3, S4, S5, S6} &
0.55 & 0.47 & 0.55 & \shortstack{S1, S2, S3, S4, S5} \\

& T2 & 0.85 & 0.59 & 0.85 & \shortstack{S1, S2} &
0.83 & 0.52 & 0.83& \shortstack{S1, S2
} &
& T2 & 0.70 & 0.42 & 0.70 & \shortstack{S1, S2, S4, S5, S6} &
0.63 & 0.52 & 0.63 & \shortstack{S1, S2, S4, S5, S6} \\

& T3 & 0.86 & 0.61 & 0.86 & \shortstack{S1, S2, S3} &
0.72 & 0.63 & 0.72 & \shortstack{S1, S2, S3} &
& T3 & 0.72 & 0.43 & 0.72 & \shortstack{S1, S2, S3, S4, S6} &
0.60 & 0.63 & 0.60 & \shortstack{S1, S2, S3, S4, S6} \\

\bottomrule
\end{tabular}}
\end{sidewaystable*}




\begin{table*}[t]
\centering
\caption{Exact sense match results across different models and time periods.}
\label{tab:Exact sense match result}

\resizebox{\textwidth}{!}{%
\begin{tabular}{c c ccccccccccc}
\toprule
\textbf{Method} & \textbf{TL} & \textbf{Eat} & \textbf{BMW} & \textbf{Brownie} & \textbf{Chronic} & \textbf{Climber} & \textbf{Germ} & \textbf{Cucumber} & \textbf{Mammy} & \textbf{Rodent} & \textbf{Salty} & \textbf{Avg} \\
\midrule

\multirow{3}{*}{\textbf{Unsupervised (ALBERT-xxlarge-v2)}} 
& T1 & 66.6\% & 50\% & 0\% & 100\% & 66.6\% & 50\% & 50\% & 66.6\% & 100\% & 33.3\% & 58.31\% \\
& T2 & 50\% & 100\% & 0\% & 50\% & 33.3\% & 0\% & 0\% & 50\% & 100\% & 16.6\% & 39.99\% \\
& T3 & 50\% & 33.3\% & 25\% & 60\% & 50\% & 50\% & 33.3\% & 75\% & 100\% & 50\% & 52.66\% \\
\midrule

\multirow{3}{*}{\textbf{N-grams and Char N-grams ML}} 
& T1 & 66.6\% & 50\% & 50\% & 66.6\% & 100\% & 80\% & 100\% & 100\% & 100\% & 83.3\% & 79.65\% \\

& T2 & 50\% & 83.3\% & 100\% & 50\% & 100\% & 100\% & 100\% & 100\% & 100\% & 83.3\% & 86.66\% \\

& T3 & 66.6\% & 66.6\% & 25\% & 100\% & 100\% & 83.3\% & 100\% & 100\% & 100\% & 66.6\% & 80.81\% \\
\midrule

\multirow{3}{*}{\textbf{DistilBERT-Based ML}} 
& T1 & 33.3\% & 100\% & 60\% & 66.6\% & 66.6\% & 75\% & 100\% & 100\% & 100\% & 66.6\% & 76.81\% \\
& T2 & 50\% & 100\% & 50\% & 75\% & 66.6\% & 75\% & 50\% & 100\% & 100\% & 50\% & 71.70\% \\
& T3 & 33.3\% & 100\% & 75\% & 60\% & 75\% & 100\% & 100\% & 75\% & 100\% & 66.6\% & 78.49\% \\
\midrule

\multirow{3}{*}{\textbf{FastText-Based ML}} 
& T1 & 50\% & 100\% & 40\% & 66.6\% & 66.6\% & 75\% & 100\% & 66.6\% & 100\% & 50\% & 71.48\% \\
& T2 & 50\% & 100\% & 50\% & 50\% & 66.6\% & 66.6\% & 100\% & 100\% & 100\% & 33.3\% & 71.65\% \\
& T3 & 50\% & 100\% & 50\% & 60\% & 50\% & 66.6\% & 100\% & 25\% & 100\% & 50\% & 65.16\% \\
\midrule

\multirow{3}{*}{\textbf{BERT-large}} 
& T1 & 83.3\% & 50\% & 60\% & 66.6\% & 100\% & 100\% & 100\% & 100\% & 100\% & 100\% & 85.99\% \\
& T2 & 50\% & 100\% & 83.3\% & 75\% & 66.6\% & 80\% & 66.6\% & 66.6\% & 100\% & 83.3\% & 77.14\% \\
& T3 & 83.3\% & 66.6\% & 87.5\% & 100\% & 75\% & 100\% & 100\% & 100\% & 100\% & 83.3\% & 89.57\% \\
\midrule

\multirow{3}{*}{\textbf{RoBERTa-large}} 
& T1 & 66.6\% & 50\% & 60\% & 66.6\% & 75\% & 100\% & 50\% & 75\% & 100\% & 83.3\% & 72.65\% \\
& T2 & 33.3\% & 100\% & 83.3\% & 75\% & 66.6\% & 75\% & 66.6\% & 66.6\% & 100\% & 83.3\% & 74.97\% \\
& T3 & 83.3\% & 66.6\% & 75\% & 100\% & 75\% & 100\% & 100\% & 100\% & 100\% & 83.3\% & 88.32\% \\
\midrule

\multirow{3}{*}{\textbf{GPT-4o-mini zero-shot}} 
& T1 & 83.3\% & 50\% & 60\% & 66.6\% & 75\% & 100\% & 66.6\% & 75\% & 100\% & 83.3\% & 75.98\% \\
& T2 & 83.3\% & 100\% & 83.3\% & 75\% & 66.6\% & 80\% & 66.6\% & 66.6\% & 100\% & 83.3\% & 80.47\% \\
& T3 & 83.3\% & 66.6\% & 75\% & 100\% & 75\% & 100\% & 66.6\% & 100\% & 100\% & 83.3\% & 84.98\% \\
\midrule

\multirow{3}{*}{\textbf{GPT-4o zero-shot}} 
& T1 & 83.3\% & 50\% & 60\% & 66.6\% & 100\% & 100\% & 100\% & 100\% & 100\% & 100\% & 85.99\% \\
& T2 & 50\% & 100\% & 83.3\% & 75\% & 66.6\% & 100\% & 66.6\% & 66.6\% & 100\% & 83.3\% & 79.14\% \\
& T3 & 83.3\% & 66.6\% & 87.5\% & 100\% & 75\% & 100\% & 100\% & 100\% & 100\% & 83.3\% & 89.57\% \\
\midrule

\multirow{3}{*}{\textbf{GPT-4o-mini few-shot}} 
& T1 & 83.3\% & 50\% & 60\% & 66.6\% & 75\% & 75\% & 66.6\% & 100\% & 100\% & 100\% & 77.65\% \\
& T2 & 83.3\% & 100\% & 83.3\% & 75\% & 66.6\% & 80\% & 66.6\% & 66.6\% & 100\% & 83.3\% & 80.47\% \\
& T3 & 66.6\% & 66.6\% & 100\% & 100\% & 75\% & 100\% & 100\% & 100\% & 100\% & 83.3\% & 89.15\% \\
\midrule

\multirow{3}{*}{\textbf{GPT-4o few-shot}} 
& T1 & 83.3\% & 50\% & 80\% & 66.6\% & 100\% & 100\% & 100\% & 100\% & 100\% & 100\% & \cellcolor{gray!20}87.99\% \\
& T2 & 83.3\% & 100\% & 83.3\% & 100\% & 100\% & 80\% & 66.6\% & 100\% & 100\% & 83.3\% & \cellcolor{gray!20}89.65\% \\
& T3 & 83.3\% & 66.6\% & 100\% & 100\% & 75\% & 100\% & 100\% & 100\% & 100\% & 83.3\% & \cellcolor{gray!20}90.82\% \\
\bottomrule
\end{tabular}%
}
\end{table*}



\begin{table*}[h]
\centering
\caption{Multi-label accuracy across different models and time periods.}
\label{tab:multi-labelled-accuracy}

\resizebox{\textwidth}{!}{%
\begin{tabular}{@{}c c ccccccccccc@{}}
\toprule
\textbf{Method} & \textbf{TP.} & \textbf{Eat} & \textbf{BMW} & \textbf{Brownie} & \textbf{Chronic} & \textbf{Climber} & \textbf{Germ} & \textbf{Cucumber} & \textbf{Mammy} & \textbf{Rodent} & \textbf{Salty} & \textbf{Avg} \\ 
\midrule

\multirow{3}{*}{\textbf{ALBERT-xxlarge-v2 (Unsupervised)}} 
& T1--T2 & 0.33 & 0.33 & 0.33 & 1 & 0.33 & 0.66 & 0 & 1 & 1 & 0.33 & 53.1\% \\ 
& T2--T3 & 0 & 0.33 & 1 & 1 & 1 & 1 & 0 & 1 & 1 & 0.33 & 66.6\% \\ 
& T1--T3 & 0 & 0.33 & 0.33 & 0.33 & 0.33 & 0.66 & 0 & 1 & 1 & 0.33 & 43.1\% \\ 
\midrule

\multirow{3}{*}{\textbf{N-grams and Char N-grams ML}} 
& T1--T2 & 0.33 & 0.33 & 1 & 0.33 & 1 & 0.66 & 1 & 1 & 1 & 0 & 66.5\% \\ 
& T2--T3 & 0.33 & 1 & 0.33 & 1 & 1 & 1 & 1 & 1 & 1 & 0.33 & \cellcolor{gray!20}79.9\% \\ 
& T1--T3 & 0 & 1 & 0.33 & 1 & 1 & 0.33 & 1 & 1 & 1 & 0 & 66.6\% \\ 
\midrule

\multirow{3}{*}{\textbf{DistilBERT-Based ML}} 
& T1--T2 & 0.33 & 1 & 0.66 & 1 & 1 & 0 & 0 & 1 & 1 & 0.33 & 63.0\% \\ 
& T2--T3 & 0.33 & 1 & 0.66 & 0.66 & 1 & 1 & 1 & 1 & 1 & 0 & 76.5\% \\ 
& T1--T3 & 1 & 1 & 1 & 1 & 1 & 1 & 1 & 0.33 & 1 & 0 & 83.3\% \\ 
\midrule

\multirow{3}{*}{\textbf{FastText-Based ML}} 
& T1--T2 & 0 & 1 & 1 & 0.33 & 0 & 0 & 1 & 0.33 & 1 & 0 & 46.6\% \\ 
& T2--T3 & 0 & 1 & 0.66 & 1 & 0.33 & 0.66 & 1 & 0.33 & 1 & 0.33 & 63.1\% \\ 
& T1--T3 & 1 & 1 & 0.66 & 1 & 0.66 & 0.66 & 1 & 0.33 & 1 & 0 & 73.1\% \\ 

\midrule

\multirow{3}{*}{\textbf{BERT-large} }
& T1--T2 & 0.33 & 0.33 & 1 & 1 & 0.33 & 1 & 0.66 & 0.33 & 1 & 0.33 & 63.1\% \\ 
& T2--T3 & 0.33 & 1 & 1 & 1 & 1 & 1 & 0.33 & 1 & 1 & 0 & 76.6\% \\ 
& T1--T3 & 1 & 1 & 1 & 1 & 0.33 & 1 & 1 & 1 & 0.33 & 0.33 & 79.9\% \\ 
\midrule

\multirow{3}{*}{\textbf{RoBERTa-large} }
& T1--T2 & 0.33 & 0.33 & 0.66 & 0.33 & 0.33 & 0 & 0.66 & 1 & 1 & 0 & 46.4\% \\ 
& T2--T3 & 0.33 & 1 & 0.66 & 1 & 1 & 1 & 0.33 & 1 & 1 & 0 & 73.2\% \\ 
& T1--T3 & 0.33 & 1 & 1 & 1 & 0.33 & 1 & 1 & 0.33 & 1 & 0 & 69.9\% \\ 
\midrule

\multirow{3}{*}{\textbf{GPT-4o-mini zero-shot}} 
& T1--T2 & 1 & 0.33 & 1 & 1 & 1 &0.66 & 0 & 1 & 1 & 0 & 69.9\% \\ 
& T2--T3 & 1 & 1 & 0.66 & 1 & 1 & 1 & 0.33 & 1 & 1 & 0 &\cellcolor{gray!20} 79.9\% \\ 
& T1--T3 & 1 & 1 & 0.66 & 1 & 1 & 1 &0.33 & 0.33 & 1 & 1 & 83.2\% \\ 
\midrule

\multirow{3}{*}{\textbf{GPT-4o zero-shot}} 
& T1--T2 & 0.33 & 0.33 & 1 & 1 & 0.33 & 1 & 0.66 & 0.33 & 1 & 0.33 & 63.1\% \\ 
& T2--T3 & 0.33 & 1 & 1 & 1 & 1 & 1 & 0.33 & 1 & 1 & 0 & 76.6\% \\ 
& T1--T3 & 1 & 1 & 1 & 1 & 0.33 & 1 & 1 & 1 & 1 & 0.33 & \cellcolor{gray!20}86.6\% \\ 
\midrule

\multirow{3}{*}{\textbf{GPT-4o-mini few-shot}} 
& T1--T2 & 1 & 0.33 & 0.66 & 1 & 1 & 0.66 & 0 & 0.33 & 1 & 0.33 & 63.1\% \\ 
& T2--T3 & 0.33 & 1 & 1 & 1 & 1 & 1 & 0.33 & 1 & 1 & 0 & 76.6\% \\ 
& T1--T3 & 0.33 & 1 & 1 & 1 & 1 & 1 & 0.33 & 1 & 1 & 0.33 & 79.9\% \\ 
\midrule

\multirow{3}{*}{\textbf{GPT-4o few-shot}} 
& T1--T2 & 1 & 0.33 & 0.66 & 1 & 1 & 0.66 & 0.66 & 1 & 1 & 0.33 & \cellcolor{gray!20}76.4\% \\ 
& T2--T3 & 1 & 1 & 1 & 1 & 0.33 & 1 & 0.33 & 1 & 1 & 0 & 76.6\% \\ 
& T1--T3 & 1 & 1 & 1 & 1 & 0.33 & 1 & 1 & 1 & 1 & 0.33 & \cellcolor{gray!20}86.6\% \\ 
\bottomrule
\end{tabular}%
}
\end{table*}



\newcolumntype{Y}{>{\RaggedRight\arraybackslash}X}

\begin{table*}[t]
\centering
\caption{Zero-shot Prompting Example.}
\label{tab:zero-shot-prompting}
\setlength{\tabcolsep}{3pt}
\renewcommand{\arraystretch}{1.25}

{\small
\begin{tabularx}{\textwidth}{@{}p{3cm}|Y@{}}
\toprule
\textbf{Prompt} & \textbf{Example} \\
\midrule

\textbf{Task explanation} &
You are tasked with analysing the word \textit{salty} in the provided text.  
Follow these steps carefully:

1. Identify the most contextually appropriate sense of \textit{salty} from these predefined meanings:

 1 — Irritated, annoyed, feeling sour, supercilious.  
 2 — Tasting of, containing, or preserved with salt.  
 3 — Old, faded, or well-used; with a positive connotation.  
 4 — Tough, aggressive, used by a veteran of a particular environment.  
 5 — Unpleasant, uncouth, crude; of language, obscene; of place.  
 6 — A name of something (e.g., a dog, company, or song). 
 
2. Write only the number of the chosen sense (1–6).  

3. Explain why this sense is the most appropriate based on the context.  \\
\midrule

\textbf{Explicit behavioural guidelines} &
Do not include “Selected Sense:”.  
Begin your reasoning immediately on the next line.  
Follow this format exactly so the system can correctly extract the reasoning. \\
\midrule

\textbf{Task instance} & 
“Ravens fans players still salty got ass kicked last week.”

\textbf{Expected Answer:} 1  

\textbf{Reasoning:} The phrase ‘still salty’ describes someone being upset or bitter about a past event.  
‘Got ass kicked’ reinforces the informal, negative emotional context. \\
\bottomrule
\end{tabularx}
}
\end{table*}



\newcolumntype{Y}{>{\RaggedRight\arraybackslash}X}

\begin{table*}[t]
\centering
\caption{Few-shot Prompting Example.}
\label{tab:few-shot-prompting}
\setlength{\tabcolsep}{3pt}
\renewcommand{\arraystretch}{1.25}

{\small
\begin{tabularx}{\textwidth}{@{}p{3cm}|Y@{}}
\toprule
\textbf{Prompt} & \textbf{Example} \\
\midrule

\textbf{Task explanation} &
You are tasked with analysing the word \textit{salty} in the provided text.
Follow these steps carefully:

1. Identify the most contextually appropriate sense of \textit{salty} from:

 1 — Irritated, annoyed, feeling sour, supercilious.  
 2 — Tasting of, containing, or preserved with salt.  
 3 — Old, faded, or well-used; with a positive connotation.  
 4 — Tough, aggressive; used by a veteran of a particular environment.  
 5 — Unpleasant, uncouth, crude; of language, obscene; of place.  
 6 — A name of something (e.g., a dog, company, or song).

2. On the first line, write only the number (1–6).  

3. On the next lines, explain why this sense is the most appropriate. \\
\midrule

\textbf{Explicit behavioural guidelines} &
Do not include “Selected Sense:”.  
Begin your reasoning immediately on the next line.  
Follow this format exactly so the system can correctly extract the reasoning. \\
\midrule

\textbf{Example instances} &

\textbf{Example 1:}  
Text: “She was salty about the team losing the championship.”  

1  

\textbf{Reasoning:} Means annoyed or upset, aligning with sense 1.

\medskip

\textbf{Example 2:}  
Text: “The soup was way too salty to eat.”  

2  

\textbf{Reasoning:} Describes taste (contains salt), aligning with sense 2.

\medskip

\textbf{Example 3:}  
Text: “His salty old jacket had seen many winters.”  

3  

\textbf{Reasoning:} Describes something old or well-used, aligning with sense 3.

\medskip

\textbf{Example 4:}  
Text: “The salty sailor shared stories from years on the sea.”  

4  

\textbf{Reasoning:} Tough, experienced sailor; aligns with sense 4.

\medskip

\textbf{Example 5:}  
Text: “The comedian’s jokes were a bit too salty for the audience.”  

5  

\textbf{Reasoning:} Refers to crude or obscene humour, aligning with sense 5. \\

\midrule

\textbf{Task instance} &

Text: “Ravens fans players still salty got ass kicked last week.”

\textbf{Expected Answer:} 1  

\textbf{Reasoning:}  
‘Still salty’ expresses annoyance or bitterness;  
‘got ass kicked’ reinforces an informal, negative emotional tone, aligning with sense 1. \\

\bottomrule
\end{tabularx}
}
\end{table*}

\clearpage
\newpage

\bibliography{sn-bibliography}

\end{document}